\documentclass[letterpaper]{article} 
\usepackage[submission]{aaai24}  
\usepackage{times}  
\usepackage{helvet}  
\usepackage{courier}  
\usepackage[hyphens]{url}  
\usepackage{graphicx} 
\usepackage{natbib}  
\usepackage{caption} 
\usepackage[hidelinks]{hyperref}
\usepackage{subcaption}
\usepackage{multirow}
\usepackage{booktabs}
\usepackage{tabularx}
\usepackage{multirow}
\usepackage{makecell}
\usepackage{framed,enumitem} 
\usepackage{amsmath,amssymb}
\usepackage{algorithm}%
\usepackage{algorithmicx}%
\usepackage{algpseudocode}%

\usepackage{array}

\newcolumntype{C}[1]{>{\centering\arraybackslash}m{#1}}
\newcolumntype{R}[1]{>{\raggedleft\arraybackslash}m{#1}}
\newcolumntype{L}[1]{>{\raggedright\arraybackslash}m{#1}}

\newcommand{\bftab}{\fontseries{b}\selectfont}

\usepackage{newfloat}
\usepackage{listings}
\DeclareCaptionStyle{ruled}{labelfont=normalfont,labelsep=colon,strut=off} 
\floatstyle{ruled}
\newfloat{listing}{tb}{lst}{}
\floatname{listing}{Listing}
\title{GradTree: Learning Axis-Aligned Decision Trees with Gradient Descent}
\author {
    Sascha Marton\textsuperscript{\rm 1},
    Stefan L\"udtke\textsuperscript{\rm 2},
    Christian Bartelt\textsuperscript{\rm 1},
    Heiner Stuckenschmidt\textsuperscript{\rm 1}
}
\affiliations {
    \textsuperscript{\rm 1} University of Mannheim, Germany \\
    \textsuperscript{\rm 2} University of Rostock, Germany\\  
    sascha.marton@uni-mannheim.de, stefan.luedtke@uni-rostock.de, christian.bartelt@uni-mannheim.de, heiner.stuckenschmidt@uni-mannheim.de
}

\begin{document}

\maketitle

\begin{abstract}
Decision Trees (DTs) are commonly used for many machine learning tasks due to their high degree of interpretability. However, learning a DT from data is a difficult optimization problem, as it is non-convex and non-differentiable. Therefore, common approaches learn DTs using a greedy growth algorithm that minimizes the impurity locally at each internal node. Unfortunately, this greedy procedure can lead to inaccurate trees.
In this paper, we present a novel approach for learning hard, axis-aligned DTs with gradient descent. The proposed method uses backpropagation with a straight-through operator on a dense DT representation, to jointly optimize all tree parameters.
Our approach outperforms existing methods on binary classification benchmarks and achieves competitive results for multi-class tasks. The method is available under: \url{https://github.com/s-marton/GradTree}.
\end{abstract}

\section{Introduction}

\begin{figure*}[ht]
\centering
\begin{subfigure}{0.5\textwidth}
  \centering
  \includegraphics[width=0.875\columnwidth]{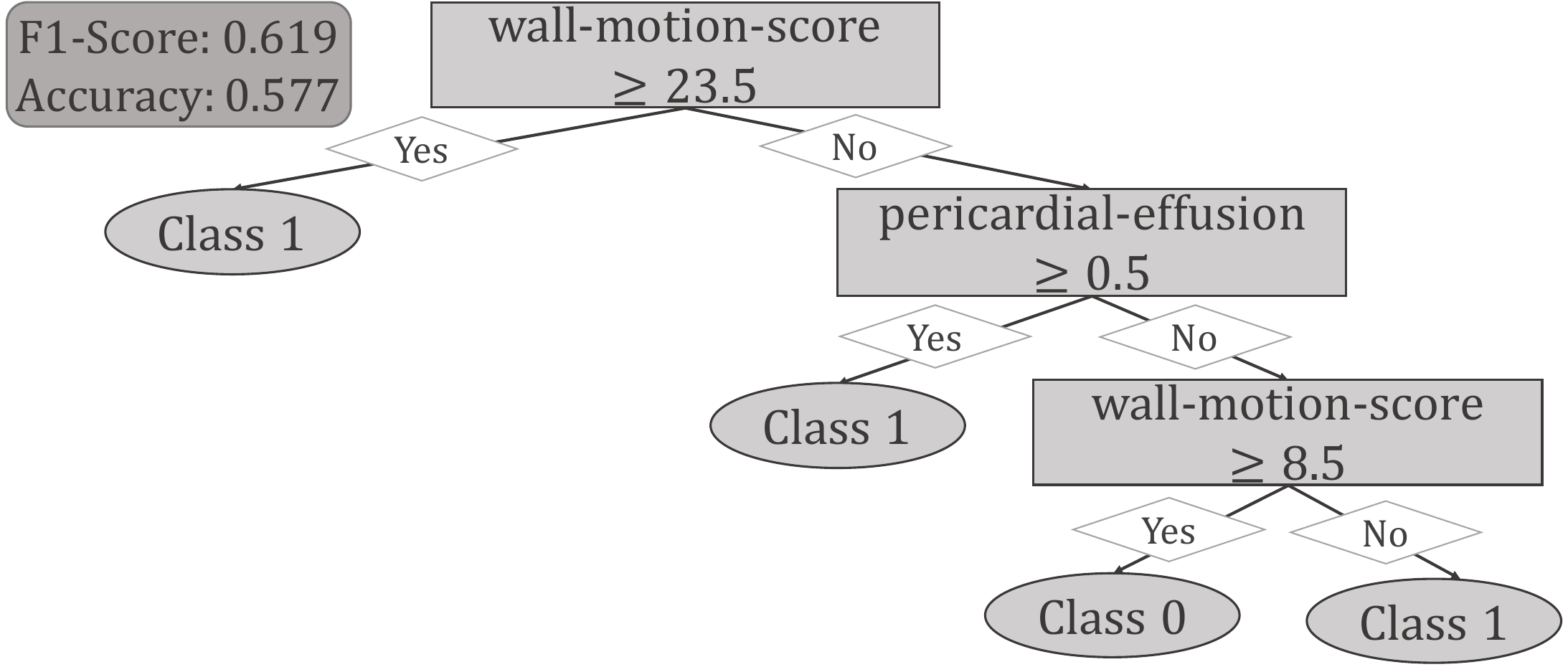}
  \caption{Greedy DT (CART)}
  \label{fig:cart_dt_example}
  
\end{subfigure}%
\begin{subfigure}{0.5\textwidth}
  \centering
  \includegraphics[width=0.875\columnwidth]{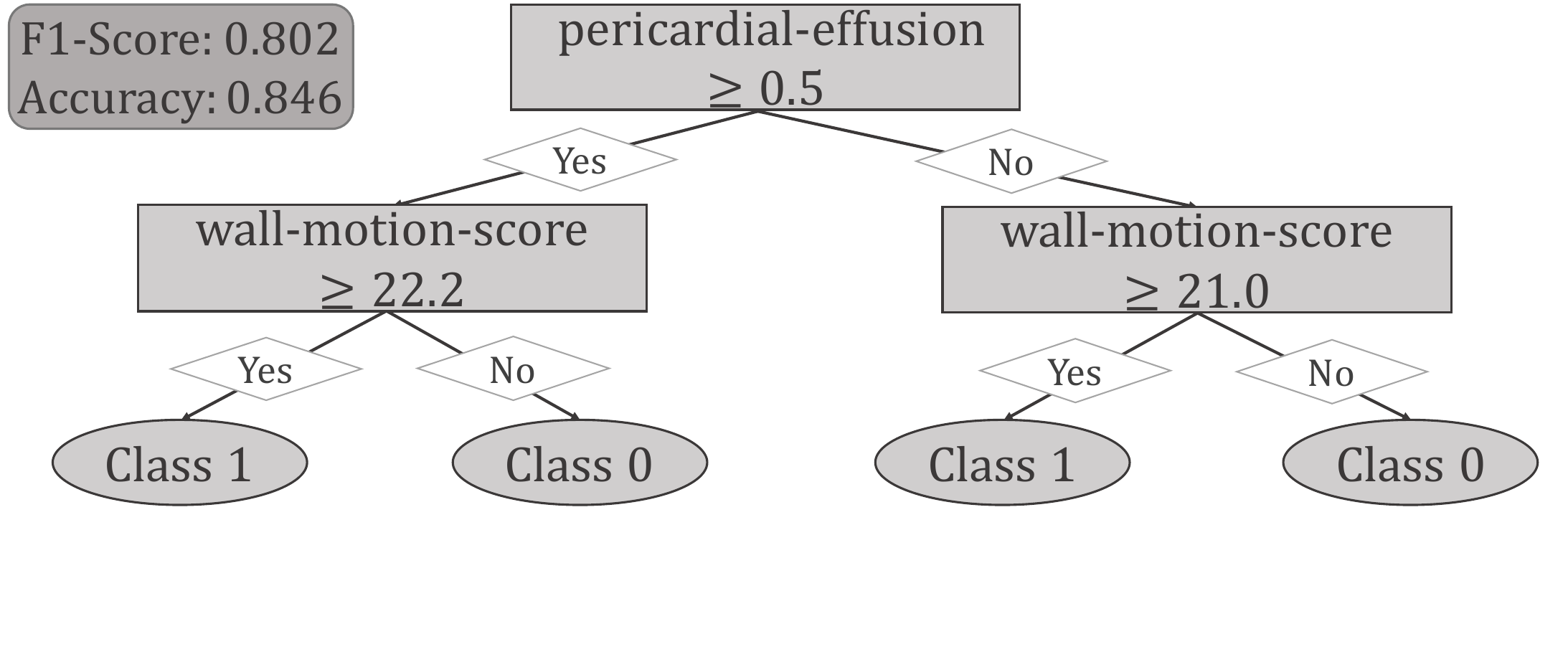}
  \caption{Gradient-Based DT (GradTree)}
  \label{fig:gdt_example}
\end{subfigure}
\caption{\textbf{Greedy vs. Gradient-Based DT.} Two DTs trained on the Echocardiogram dataset. 
The CART DT (left) makes only locally optimal splits, while GradTree (right) jointly optimizes all parameters, leading to significantly better performance.
}
\label{fig:dt_comparison_example}
\end{figure*}

Decision trees (DTs) are some of the most popular machine learning models and are still frequently used today. In particular, with the growing interest in explainable artificial intelligence (XAI), DTs have regained popularity due to their interpretability.
However, learning a DT is a difficult optimization problem, since it is non-convex and non-differentiable. 
Therefore, the prevailing approach to learn a DT is a greedy procedure that minimizes the impurity at each internal node. The algorithms still in use today, such as CART~\citep{cart_breiman1984} and C4.5~\citep{quinlan1993c45}, were developed in the 1980s and have remained largely unchanged since then. 
Unfortunately, greedy algorithms optimize the objective locally at each internal node which constrains the search space, and potentially leads to inaccurate trees, as illustrated below:

\emph{Example 1}
The Echocardiogram dataset~\citep{ml_repository} deals with predicting one-year survival of patients after a heart attack based on tabular data from an echocardiogram. Figure~\ref{fig:dt_comparison_example} shows two DTs.
The tree on the left is learned by a greedy algorithm (CART) while the one on the right is learned with our gradient-based approach.
We can observe that the greedy procedure leads to a tree with a significantly lower performance. Splitting on the \emph{wall-motion-score} is the locally optimal split (see Figure~\ref{fig:cart_dt_example}), but globally, it is beneficial to split based on the \emph{wall-motion-score} with different values conditioned on the \emph{pericardial-effusion} in the second level (Figure~\ref{fig:gdt_example}). 

In this paper, we propose a novel approach for learning hard, axis-aligned DTs based on a joint optimization of all tree parameters using gradient descent, which we call \textbf{Grad}ient-Based Decision \textbf{Tree} (GradTree). 
Similar to optimization in neural networks, GradTree yields a desirable local optimum of parameters that generalizes well to test data. 
Using a gradient-based optimization, GradTree can overcome the limitations of greedy approaches, which are constrained by sequentially selecting optimal splits, as illustrated in Figure~\ref{fig:dt_comparison_example}.
At the same time, GradTree can converge to a local optimum that offers good generalization, and thus provides an advantage over alternative non-greedy methods like optimal DTs~\citep{murtree,aglin2020learning}, which often suffer from severe overfitting~\citep{zantedeschi2021learning}.

Specifically, our contributions are: 
\begin{itemize}
    \item We introduce a dense DT representation that enables a joint, gradient-based optimization of all tree parameters (Section~\ref{ssec:dense_rep}).
    \item We present a procedure to deal with the non-differentiable nature of DTs using backpropagation with a straight-through (ST) operator (Section~\ref{ssec:adjusted_backprop}).
    \item We propose a novel tree routing that allows an efficient, parallel optimization of all tree parameters with gradient descent (Section~\ref{ssec:training}).
\end{itemize}

We empirically evaluate GradTree on a large number of real-world datasets (Section~\ref{sec:evaluation}). 
GradTree outperforms existing methods for binary classification tasks and achieves competitive results on multi-class datasets.
On several benchmark datasets, the performance difference between GradTree and other methods is substantial. 
The gradient-based optimization of GradTree also provides more flexibility by allowing split adjustments during training and easy integration of custom loss functions.

\section{Related Work}
\label{sec:relatedwork}
\paragraph{Greedy DT Algorithms} 
The most prominent DT learning algorithms still frequently used, namely CART~\citep{cart_breiman1984} and C4.5~\citep{quinlan1993c45}, date back to the 1980s.
Both follow a greedy procedure to learn a DT. 
Since then, many variations to those algorithms have been proposed, for instance C5.0~\citep{kuhn2013applied_c50} and GUIDE~\citep{loh2002regression_guide1,loh2009improving_guide2}.
However, until today, none of these algorithms was able to consistently outperform CART and C4.5 as shown for instance by \citet{zharmagambetov2021non}. 

\paragraph{Optimal DTs} To overcome the issues of a greedy DT induction, many researchers focused on finding an efficient alternative. Optimal DTs aim to optimize an objective (e.g., the purity) through an approximate brute force search to find a globally optimal tree with a certain specification~\citep{zharmagambetov2021non}. 
Therefore, they most commonly use mixed integer optimization~\citep{bertsimas2017optimal} or a branch-and-bound algorithm to remove irrelevant parts from the search space~\citep{aglin2020learning,lin2020generalized}.
MurTree~\citep{murtree} further uses dynamic programming, which reduces the runtime significantly. 
However, most state-of-the-art approaches still require binary data and therefore a discretization of continuous features~\citep{bertsimas2017optimal,aglin2020learning,murtree}, which can lead to information loss. An exception is the approach by \citet{mazumder2022quant}, which can handle continuous features out-of-the-box. However, their method is optimized for very sparse trees and limited to a maximum depth of $3$.

While optimal DTs search for a global optimum, GradTree does not necessarily pursue this. Instead, like optimization in neural networks, it aims for a local optimum that offers good generalization to test data.

We further want to emphasize that the local optima that can be reached by GradTree have a significant advantage over the local optimum of a greedy approach: While the local optimum of greedy approaches is constrained by sequentially selecting the optimal split at each node, GradTree overcomes this limitation by optimizing all parameters jointly.

\paragraph{Genetic DTs} Another way to learn DTs in a non-greedy fashion is by using evolutionary algorithms. Evolutionary algorithms perform a robust global search in the space of candidate solutions based on the concept of survival of the fittest~\citep{barros2011survey}. This usually results in smaller trees and a better identification of feature interactions compared to a greedy, local search~\citep{freitas2002data}. 

\paragraph{Oblique DTs} In contrast to vanilla DTs that make a hard decision at each internal node, many hierarchical mixture of expert models~\citep{jordan1994hierarchical} have been proposed. They usually make soft splits, where each branch is associated with a probability~\citep{irsoy2012soft,frosst2017distilling}. Further, the models do not comprise univariate, axis-aligned splits, but are oblique with respect to the axes. 
These adjustments to the tree architecture allow for the application of further optimization algorithms, including gradient descent. 
\citet{blanquero2020sparsity} aim to increase the interpretability of oblique trees by optimizing for sparsity, using fewer variables at each split and simultaneously fewer splits in the whole tree.
\citet{tanno2019adaptive} combine the benefits of neural networks and DTs, using so-called adaptive neural trees (ANTs). They employ a stochastic routing based on a Bernoulli distribution and utilize non-linear transformer modules at the edges, making the resulting trees soft and oblique.
\citet{one-stage-tree} propose One-Stage Tree as a novel method for learning soft DTs, including the tree structure, while maintaining discretization during training, which results in a higher interpretability compared to existing soft DTs. However, in contrast to GradTree, the routing is instance-wise, which significantly hampers a global interpretation of the model.
\citet{norouzi2015efficient} proposed an approach to overcome the need for soft decisions to apply gradient-based algorithms by minimizing a convex-concave upper bound on the tree's empirical loss. While this allows the use of hard splits, the approach is still limited to oblique trees. 
\citet{zantedeschi2021learning} use argmin differentiation to simultaneously learn all tree parameters by relaxing a mixed-integer program for discrete parameters to allow for gradient-based optimization. This allows hard splits, but in contrast to GradTree, they still require a differentiable split function (e.g., a linear function which results in oblique trees).
Similarly, \citet{karthikeyanlearning} developed a gradient-based approach to learn hard DTs. Like to GradTree, they use an ST operator to handle the hard step functions. While their formulation is limited to oblique trees, our approach permits axis-aligned DTs.

In summary, unlike oblique DTs, GradTree allows hard, axis-aligned splits that consider only a single feature at each split, providing significantly higher interpretability, especially at the split-level. This is supported by \citet{molnar2020interpretable} where the authors argue that humans cannot comprehend explanations involving more than three dimensions at once.

\paragraph{Oblivious DT Ensembles} \citet{neural_oblivious} proposed an oblivious tree ensemble for deep learning. Oblivious DTs use the same splitting feature and threshold in all internal nodes of the same depth, making them only suitable as weak learners in an ensemble. They use an entmax transformation of the choice function and a two-class entmax as split function. 
This results in oblique and soft trees, while GradTree is axis-aligned and hard.
\citet{node_gam} proposed a temperature annealing procedure to gradually turn the input to an entmax function one-hot, which can enforce axis-aligned trees. 
In contrast, our approach employs an ST operator immediately following an entmax transformation to yield a one-hot encoded vector. Our experiments substantiate that our method achieves superior results in the context of individual DTs.

\paragraph{Axis-Aligned DT Ensembles} While existing methods for learning tree ensembles often require structural changes, such as relaxing axis-aligned splits to oblivious ones, \citet{bruch2020learning} use input perturbation as a purely analytical procedure to approximate the gradients of the ensemble. As a result, similar to GradTree, it is possible to learn and fine-tune an axis-aligned DT ensemble. Although their method makes this feasible, the authors decided not to learn the split index but only the split threshold, in order to avoid potential changes to the model's structure. In contrast, GradTree uses a dense representation that allows for the efficient learning of both the split threshold and the feature index to split on. This is particularly important when aiming for an interpretable DT model, as opposed to a large ensemble, since the number of possible splits is limited.

\paragraph{Deep Neural Decision Trees (DNDTs)} \citet{dndt} propose DNDTs that realize tree models as neural networks, utilizing a soft binning function for splitting. Therefore, the resulting trees are soft, but axis-aligned, which makes this work closely related to our approach.
Since DNDTs are generated via the Kronecker product of the binning layers, the structure depends on the number of features and classes (and the number of bins). As discussed by the authors, this results in  poor scalability w.r.t.\ the number of features, which currently can only be solved by using random forests for high-dimensional datasets ($>12$ features). Our approach, in contrast, scales linearly with the number of features, making it efficient for high-dimensional datasets. Furthermore, using the Kronecker product to build the tree prevents splitting on the same feature with different thresholds in the same path, which can be crucial to achieve a good performance. 
For GradTree, both the split threshold and the split index are learned parameters, inherently allowing the model to split on the same feature multiple times.

\section{GradTree: Gradient-Based Decision Trees}\label{sec:gradient_dts}
In this section, we introduce GradTree. 
We present a new DT representation and a novel algorithm that allows learning hard, axis-aligned DTs with gradient descent. 
More specifically, we will use backpropagation with a straight-through (ST) operator (Section~\ref{ssec:adjusted_backprop}) on a dense DT representation (Section~\ref{ssec:dense_rep}) to adjust the model parameters during the training. Furthermore, our novel tree routing (Section~\ref{ssec:training}) allows an efficient optimization of all parameters over an entire batch with a single set of matrix operations.

\begin{figure*}[tb]
\centering
\begin{subfigure}{0.5\textwidth}
   \centering
  \includegraphics[width=0.9\columnwidth]{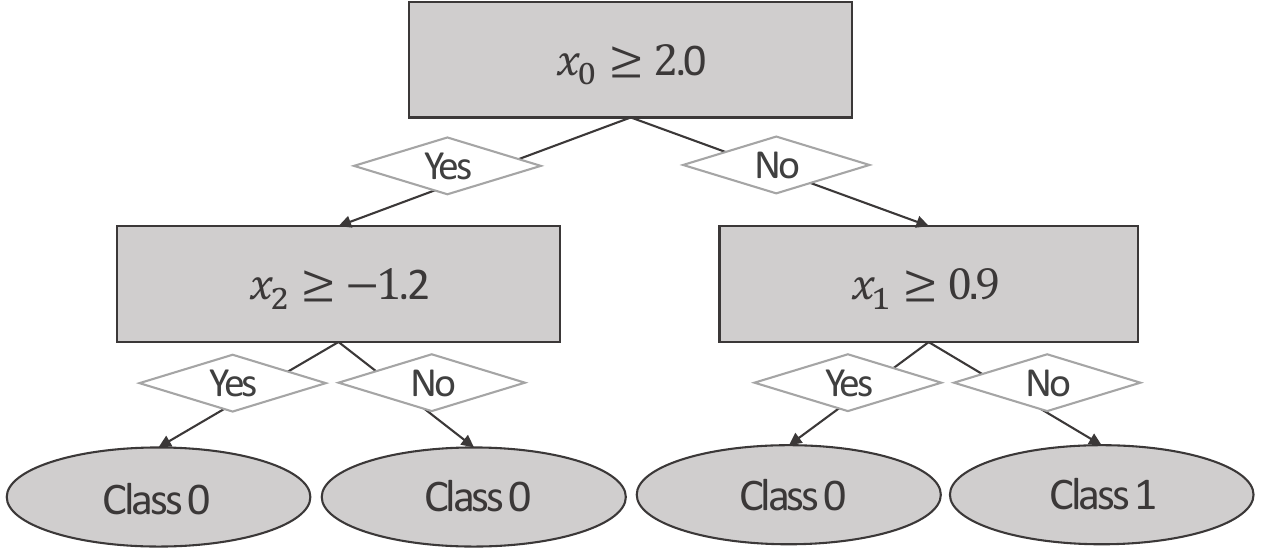}
  \caption{Vanilla DT Representation}
  \label{fig:old_rep_2}
  
\end{subfigure}%
\begin{subfigure}{0.5\textwidth}
  \centering
  \includegraphics[width=0.9\columnwidth]{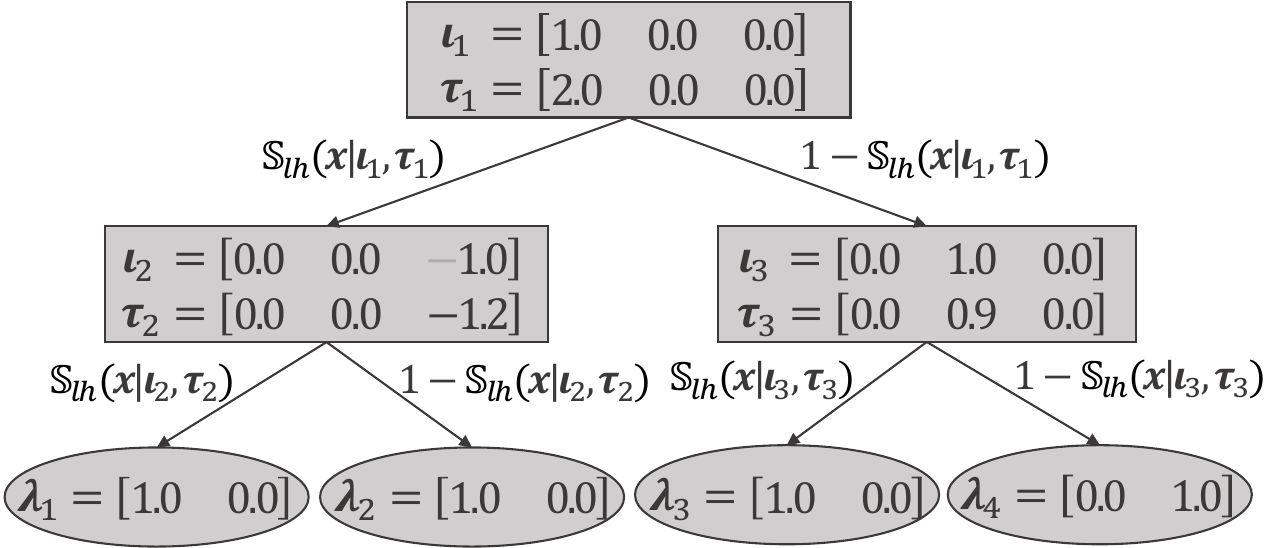}
  \caption{Dense DT Representation}
  \label{fig:new_rep_dense}
\end{subfigure}
\caption{\textbf{Standard vs. Dense DT Representation.} Comparison of a standard and the equivalent dense representation for an exemplary DT with depth $2$ and a dataset with $3$ variables and $2$ classes. Here, $\mathbb{S}_{\text{lh}}$ stands for $\mathbb{S}_\text{logistic\_hard}$ (Equation~\ref{eq:split_sigmoid_round}). }
\label{fig:dt_reps_dense}
\end{figure*}
\subsection{Arithmetic Decision Tree Formulation}
Here, we introduce a notation for DTs with respect to their parameters. We formulate DTs as an arithmetic function based on addition and multiplication, rather than as a nested concatenation of rules, which is necessary for a gradient-based learning. 
Note that our notation and training procedure assume fully-grown (i.e. complete, full) DTs. After training, we apply a basic post-hoc pruning to reduce the tree size for application.
Our formulation aligns with \citet{kontschieder2015deep}. 
However, they only consider stochastic routing and oblique trees, whereas our formulation emphasizes deterministic routing and axis-aligned trees.

For a DT of depth $d$, the parameters include one split threshold and one feature index for each internal node, represented as vectors \(\boldsymbol{\tau} \in \mathbb{R}^{2^d-1}\) and \(\boldsymbol{\iota} \in \mathbb{N}^{2^d-1}\) respectively, where \(2^d-1\) equals the number of internal nodes in a fully-grown DT.
Additionally, each leaf node comprises a class membership, in the case of a classification task, which we denote as the vector $\boldsymbol{\lambda} \in \mathcal{C}^{2^d}$, where $\mathcal{C}$ is the set of classes and $2^d$ equals the number of leaf nodes in a fully-grown DT.

Formally, a DT can be expressed as a function $ DT(\cdot | \boldsymbol{\tau}, \boldsymbol{\iota}, \boldsymbol{\lambda}): \mathbb{R}^n \rightarrow \mathcal{C}$ with respect to its parameters:
\begin{equation}\label{eq:dt_prediction}
\small
    DT(\boldsymbol{x} | \boldsymbol{\tau}, \boldsymbol{\iota}, \boldsymbol{\lambda}) = \sum_{l=0}^{2^d-1} \lambda_l \, \mathbb{L}(\boldsymbol{x} | l, \boldsymbol{\tau}, \boldsymbol{\iota}) 
\end{equation}
The function $\mathbb{L}$ indicates whether a sample $\boldsymbol{x} \in \mathbb{R}^n$ belongs to a leaf $l$, and can be defined as a multiplication of the split functions of the preceding internal nodes. We define the split function $\mathbb{S}$ as a Heaviside step function
\begin{equation}\label{eq:split_heaviside}
\small
    \mathbb{S}_{\text{Heaviside}}(\boldsymbol{x} | \iota, \tau) =
    \begin{cases}
        1, \text{if } x_\iota \ge \tau \\
        0, \text{otherwise}
    \end{cases}
\end{equation}
where $\iota$ is the index of the feature considered at a certain split and $\tau$ is the corresponding threshold.

By enumerating the internal nodes of a fully-grown tree with depth $d$ in a breadth-first order, we can now define the indicator function $\mathbb{L}$ for a leaf $l$ as

\begin{equation}\label{eq:indicator_func}
\small
\begin{split}
    \mathbb{L}(\boldsymbol{x} | l, \boldsymbol{\tau}, \boldsymbol{\iota}) = \prod^d_{j=1} & \left(1 - \mathfrak{p}(l,j) \right) \, \mathbb{S}(\boldsymbol{x} | \tau_{\mathfrak{i}(l,j)}, \iota_{\mathfrak{i}(l,j)}) \\
    + & \mathfrak{p}(l,j) \, \left( 1 - \mathbb{S}(\boldsymbol{x} | \tau_{\mathfrak{i}(l,j)}, \iota_{\mathfrak{i}(l,j)}) \right)
\end{split}
\end{equation}

Here, $\mathfrak{i}$ is the index of the internal node preceding a leaf node $l$ at a certain depth $j$ and can be calculated as

\begin{equation}\label{eq:index_calc}
\small
    \mathfrak{i}(l,j) = 2^{j-1} + \left\lfloor \frac{l}{2^{d-(j-1)}} \right\rfloor - 1 
\end{equation}

Additionally, $\mathfrak{p}$ indicates whether the left ($\mathfrak{p} = 0$) or the right branch ($\mathfrak{p} = 1$) was taken at the internal node preceding a leaf node $l$ at a certain depth $j$. We can calculate $\mathfrak{p}$ as

\begin{equation}\label{eq:path_calc}
\small
    \mathfrak{p}(l,j) = \left\lfloor \frac{l}{2^{d-j}} \right\rfloor \bmod 2
\end{equation}

As becomes evident, DTs involve non-differentiable operations in terms of the split function, including the split feature selection (Equation~\ref{eq:split_heaviside}). This precludes the application of backpropagation for learning the parameters. 
Specifically, to efficiently learn a DT using backpropagation, we must address three challenges:
\begin{itemize}
    \item[] \textbf{C1} The index $\iota$ for the split feature selection is defined as $\iota \in \mathbb{N}$. However, the index $\iota$ is a parameter of the DT and a gradient-based optimization requires $\iota \in \mathbb{R}$.    
    \item[] \textbf{C2} The split function $\mathbb{S}(\boldsymbol{x}| \iota, \tau)$ is a Heaviside step function with an undefined gradient for $x_\iota = \tau$ and $0$ gradient elsewhere, which precludes an efficient optimization. 
    \item[] \textbf{C2} Leafs in a vanilla DT comprise a class membership $\lambda \in \mathcal{C}$. To calculate an informative loss and optimize the leaf parameters with gradient descent, we need $\lambda \in \mathbb{R}$.       
\end{itemize}

Additionally, the computation of the internal node index $\mathfrak{i}$ and path position $\mathfrak{p}$ involves non-differentiable operations. However, given our focus on fully-grown trees, these values remain constant, allowing for their computation prior to the optimization process.

\subsection{Dense Decision Tree Representation}\label{ssec:dense_rep}
In this subsection, we present a differentiable representation of the feature indices $\boldsymbol{\iota}$ to facilitate gradient-based optimization, which is illustrated in Figure~\ref{fig:dt_reps_dense}.

To this end, we expand the vector $\boldsymbol{\iota} \in \mathbb{R}^{2^d-1}$ to a matrix $I \in \mathbb{R}^{2^d-1} \times \mathbb{R}^{n}$. This is achieved by one-hot encoding the feature index as $\boldsymbol{\iota} \in \mathbb{R}^{n}$ for each internal node.
This adjustment is necessary for the optimization process to account for the fact that feature indices are categorical instead of ordinal. 
Although our matrix representation for feature selection has parallels with that proposed by \citet{neural_oblivious}, we introduce a novel aspect: A matrix representation for split thresholds. We denote this representation as $T \in \mathbb{R}^{2^d-1} \times \mathbb{R}^{n}$. Instead of representing a single value for all features, we store individual values for each feature, denoted as $\boldsymbol{\tau} \in \mathbb{R}^{n}$.
This modification is tailored to support the optimization process, particularly in recognizing that split thresholds are feature-specific and non-interchangeable. In essence, a viable split threshold for one feature may not be suitable for another. This adjustment acts as a memory mechanism, ensuring that a given split threshold is exclusively associated with the corresponding feature. 
Consequently, this refinement enhances the exploration of feature selection at every split during the training.
Besides the previously mentioned advantages, using a dense DT representation allows the use of matrix multiplications for an efficient computation.
Accordingly, we can reformulate the Heaviside split function (Equation~\ref{eq:split_heaviside}) as

\begin{align}\label{eq:split_sigmoid}
\small
     & \mathbb{S}_{\text{logistic}}(\boldsymbol{x}| \boldsymbol{\iota}, \boldsymbol{\tau}) =  S \left( \sum_{i=0}^n \iota_i  x_i - \sum_{i=0}^n \iota_i  \tau_i \right)  \\\label{eq:split_sigmoid_round}
     & \mathbb{S}_{\text{logistic\_hard}}(\boldsymbol{x}| \boldsymbol{\iota}, \boldsymbol{\tau}) =  \left\lfloor \mathbb{S}_{\text{logistic}}(\boldsymbol{x}| \boldsymbol{\iota}, \boldsymbol{\tau}) \right\rceil 
\end{align}

where \(S(x)=\frac{1}{1+e^{-x}}\) denotes the logistic function and \(\lfloor \cdot \rceil\) represents for rounding to the nearest integer.
In our context, with $\boldsymbol{\iota}$ being one-hot encoded, $\mathbb{S}_{\text{logistic\_hard}}(\boldsymbol{x}| \boldsymbol{\iota}, \boldsymbol{\tau}) = \mathbb{S}_{\text{Heaviside}}(\boldsymbol{x}| \iota, \tau)$ holds.

\subsection{Backpropagation of Decision Tree Loss}\label{ssec:adjusted_backprop}
While the dense representation introduced previously emphasizes an efficient learning of axis-aligned DTs, it does not solve \textbf{C1}-\textbf{C3}. 
In this subsection, we will address those challenges by using the ST operator for backpropagation. 

For the function value calculation in the forward pass, we need to assure that $\boldsymbol{\iota}$ is a one-hot encoded vector. This can be achieved by applying a hardmax function on the feature index vector for each internal node. However, applying a hardmax is a non-differentiable operation, which precludes gradient computation. To overcome this issue, we use the ST operator~\citep{bengio2013propagating}: For the forward pass, we apply the hardmax as is. For the backward pass, we exclude this operation and directly propagate back the gradients of $\boldsymbol{\iota}$. Accordingly, we can optimize the parameters of $\boldsymbol{\iota}$ where $\boldsymbol{\iota} \in \mathbb{R}$ while still using axis-aligned splits during training (\textbf{C1}).
However, this procedure introduces a mismatch between the forward and backward pass. To reduce this mismatch, we additionally perform an entmax transformation~\citep{entmax_transformation} to generate a sparse distribution over $\boldsymbol{\iota}$ before applying the hardmax. 

Similarly, we employ the ST operator to ensure hard splits (Equation~\ref{eq:split_sigmoid_round}) by excluding $\lfloor \cdot \rceil$ for the backward pass (\textbf{C2}). Using the sigmoid logistic function before applying the ST operator (see Equation~\ref{eq:split_sigmoid}) utilizes the distance to the split threshold as additional information for the gradient calculation. If the feature considered at an internal node is close to the split threshold  for a specific sample (i.e. it is more important for the current decision), this will result in larger gradients compared to a sample that is more distant.

Furthermore, we need to adjust the leaf nodes of the DT to allow an efficient loss calculation (\textbf{C3}). Vanilla DTs contain the predicted class for each leaf node and are functions $DT: \mathbb{R}^n \rightarrow \mathcal{C}$. We use a probability distribution at each leaf node and therefore define DTs as a function $DT: \mathbb{R}^n \rightarrow \mathbb{R}^c$ where $c$ is the number of classes. Consequently, the parameters of the leaf nodes are defined as $L \in \mathbb{R}^{2^d} \times \mathbb{R}^c$ for the whole tree and $\boldsymbol{\lambda} \in \mathbb{R}^{c}$ for a specific leaf node. This adjustment allows the application of standard loss functions.

\subsection{Deterministic Tree Routing and Training}\label{ssec:training}
In the previous subsections, we introduced the adjustments that are necessary to apply gradient descent to DTs.
During the optimization, we calculate the gradients with backpropagation based on the computation graph of the tree pass function.
The tree pass function to calculate the function values is summarized in Algorithm~\ref{alg:forward} and utilizes the adjustments introduced in the previous sections. 
Our tree routing facilitates the computation of the tree pass function over a complete batch as a single set of matrix operations, which allows an efficient computation.
We also want to note that our dense representation can be converted into an equivalent vanilla DT representation at each point in time. Similarly, the fully-grown nature of GradTree is only required during the gradient-based optimization and standard pruning techniques to reduce the tree size are applied post-hoc.

Furthermore, our implementation optimizes the gradient descent algorithm by leveraging common stochastic gradient descent techniques, including mini-batch calculation and momentum using the Adam optimizer~\citep{kingma2014adam} with weight averaging~\citep{izmailov2018averaging}. 
Moreover, we implement early stopping and random restarts to avoid bad initial parametrizations, where the best parameters are selected based on the validation loss. Further details can be found in Appendix~\ref{A:hyperparams}.

\begin{algorithm}[tb]
\small
\caption{Tree Pass Function}   \label{alg:forward} 

\begin{algorithmic}[1] 
    
    \Function{pass}{$I, T, L, \boldsymbol{x}$}
        \State{$I \leftarrow \text{entmax}(I)$} 
        \State{$I \leftarrow I - c^*_1 \quad \text{where } c^*_1 = I - \text{hardmax}(I)$} \Comment{ST operator}
        
        \State{$\hat{\boldsymbol{y}} \leftarrow {[0]}^c$}
        \For{$l = 0, \dots, 2^d-1$}
            \State{$p \leftarrow 1$}
            \For{$j = 1, \dots, d$}
                \State{$ \mathfrak{i} \leftarrow 2^{j-1} + \left\lfloor \frac{l}{2^{d-(j-1)}} \right\rfloor - 1 $} \Comment{Equation~\ref{eq:index_calc}}            
            
                \State{$\mathfrak{p} \leftarrow \left\lfloor \frac{l}{2^{d-j}} \right\rfloor \bmod 2$} \Comment{Equation~\ref{eq:path_calc}}
                

                \State{$s \leftarrow S\left(\sum\limits^n_{i=0} T_{\mathfrak{i}, i} \, I_{\mathfrak{i}, i} - \sum\limits^n_{i=0} x_{i} \, I_{\mathfrak{i}, i}\right)$} \Comment{Equation~\ref{eq:split_sigmoid}}

                
                \State{$s \leftarrow s - c^*_2 \quad \text{where } c^*_2 = s - \lfloor s \rceil$} \Comment{ST operator}

                \State{$p \leftarrow p \, \left( (1-\mathfrak{p}) \, s + \mathfrak{p} \, (1 - s) \right)$} \Comment{Equation~\ref{eq:indicator_func}}
            \EndFor
            \State{$\hat{\boldsymbol{y}} \leftarrow \hat{\boldsymbol{y}} + L_l \, p$} \Comment{Equation~\ref{eq:dt_prediction}}
        \EndFor
        \State \Return{$\sigma \left( \hat{\boldsymbol{y}} \right)$} \Comment{Softmax  $\sigma$ to get probability distribution}
    \EndFunction
\end{algorithmic}

\end{algorithm}

\section{Experimental Evaluation}\label{sec:evaluation}

The following experiments aim to evaluate the predictive capability of GradTree against existing methods.

\subsection{Experimental Setup} 

\begin{table*}[tb]
\centering
\small
\resizebox{0.9\textwidth}{!}{
\begin{tabular}{lccccc}
\toprule
{} &                           \multicolumn{2}{l}{Gradient-Based} &            \multicolumn{2}{l}{Non-Greedy} &                       \multicolumn{1}{l}{Greedy} \\ \cmidrule(lr){2-3}  \cmidrule(lr){4-5} \cmidrule(lr){6-6}
  
{} &                           \multicolumn{1}{l}{GradTree (ours)} &                   \multicolumn{1}{l}{DNDT} &            \multicolumn{1}{l}{GeneticTree} &                          \multicolumn{1}{l}{DL8.5 (Optimal)} &                       \multicolumn{1}{l}{CART} \\
\midrule
Blood Transfusion                  &  \bftab 0.628 $\pm$ .036 (1) &         0.543 $\pm$ .051 (5) &         0.575 $\pm$ .094 (4) &  0.590 $\pm$ .034 (3) &         0.613 $\pm$ .044 (2) \\
Banknote Authentication          &  \bftab 0.987 $\pm$ .007 (1) &         0.888 $\pm$ .013 (5) &         0.922 $\pm$ .021 (4) &  0.962 $\pm$ .011 (3) &         0.982 $\pm$ .007 (2) \\ 
Titanic                            &  \bftab 0.776 $\pm$ .025 (1) &         0.726 $\pm$ .049 (5) &         0.730 $\pm$ .074 (4) &  0.754 $\pm$ .031 (2) &         0.738 $\pm$ .057 (3) \\
Raisins                            &         0.840 $\pm$ .022 (4) &         0.821 $\pm$ .033 (5) &  \bftab 0.857 $\pm$ .021 (1) &  0.849 $\pm$ .027 (3) &         0.852 $\pm$ .017 (2) \\
Rice                               &         0.926 $\pm$ .007 (3) &         0.919 $\pm$ .012 (5) &         0.927 $\pm$ .005 (2) &  0.925 $\pm$ .008 (4) &  \bftab 0.927 $\pm$ .006 (1) \\
Echocardiogram                     &  \bftab 0.658 $\pm$ .113 (1) &         0.622 $\pm$ .114 (3) &         0.628 $\pm$ .105 (2) &  0.609 $\pm$ .112 (4) &         0.555 $\pm$ .111 (5) \\
Wisconcin Breast Cancer &         0.904 $\pm$ .022 (2) &  \bftab 0.913 $\pm$ .032 (1) &         0.892 $\pm$ .028 (4) &  0.896 $\pm$ .021 (3) &         0.886 $\pm$ .025 (5) \\ 
Loan House                         &  \bftab 0.714 $\pm$ .041 (1) &         0.694 $\pm$ .036 (2) &         0.451 $\pm$ .086 (5) &  0.607 $\pm$ .045 (4) &         0.662 $\pm$ .034 (3) \\
Heart Failure                      &         0.750 $\pm$ .070 (3) &         0.754 $\pm$ .062 (2) &         0.748 $\pm$ .068 (4) &  0.692 $\pm$ .062 (5) &  \bftab 0.775 $\pm$ .054 (1) \\
Heart Disease                      &  \bftab 0.779 $\pm$ .047 (1) &             $n>12$ &         0.704 $\pm$ .059 (4) &  0.722 $\pm$ .065 (2) &         0.715 $\pm$ .062 (3) \\
Adult                              &         0.743 $\pm$ .034 (2) &             $n>12$ &         0.464 $\pm$ .055 (4) &  0.723 $\pm$ .011 (3) &  \bftab 0.771 $\pm$ .011 (1) \\
Bank Marketing                     &  \bftab 0.640 $\pm$ .027 (1) &             $n>12$ &         0.473 $\pm$ .002 (4) &  0.502 $\pm$ .011 (3) &         0.608 $\pm$ .018 (2) \\
Congressional Voting               &  \bftab 0.950 $\pm$ .021 (1) &             $n>12$ &         0.942 $\pm$ .021 (2) &  0.924 $\pm$ .043 (4) &         0.933 $\pm$ .032 (3) \\ 
Absenteeism                        &  \bftab 0.626 $\pm$ .047 (1) &             $n>12$ &         0.432 $\pm$ .073 (4) &  0.587 $\pm$ .047 (2) &         0.564 $\pm$ .042 (3) \\
Hepatitis                          &         0.608 $\pm$ .078 (2) &             $n>12$ &         0.446 $\pm$ .024 (4) &  0.586 $\pm$ .083 (3) &  \bftab 0.622 $\pm$ .078 (1) \\
German                             &  \bftab 0.592 $\pm$ .068 (1) &             $n>12$ &         0.412 $\pm$ .006 (4) &  0.556 $\pm$ .035 (3) &         0.589 $\pm$ .065 (2) \\
Mushroom                           &  \bftab 1.000 $\pm$ .001 (1) &             $n>12$ &         0.984 $\pm$ .003 (4) &  0.999 $\pm$ .001 (2) &         0.999 $\pm$ .001 (3) \\
Credit Card                        &         0.674 $\pm$ .014 (4) &             $n>12$ &  \bftab 0.685 $\pm$ .004 (1) &  0.679 $\pm$ .007 (3) &         0.683 $\pm$ .010 (2) \\
Horse Colic                        &  \bftab 0.842 $\pm$ .039 (1) &             $n>12$ &         0.496 $\pm$ .169 (4) &  0.708 $\pm$ .038 (3) &         0.786 $\pm$ .062 (2) \\
Thyroid                            &         0.905 $\pm$ .010 (2) &             $n>12$ &         0.605 $\pm$ .116 (4) &  0.682 $\pm$ .018 (3) &  \bftab 0.922 $\pm$ .011 (1) \\
Cervical Cancer                    &  \bftab 0.521 $\pm$ .043 (1) &             $n>12$ &         0.514 $\pm$ .034 (2) &  0.488 $\pm$ .027 (4) &         0.506 $\pm$ .034 (3) \\
Spambase                           &         0.903 $\pm$ .025 (2) &             $n>12$ &         0.863 $\pm$ .019 (3) &  0.863 $\pm$ .011 (4) &  \bftab 0.917 $\pm$ .011 (1) \\
\midrule
Mean Reciprocal Rank (MRR) $\uparrow$                    &  \bftab 0.758 $\pm$ .306 (1) &         0.370 $\pm$ .268 (3) &         0.365 $\pm$ .228 (4) &  0.335 $\pm$ .090 (5) &         0.556 $\pm$ .293 (2) \\
Mean Relative Diff. (MRD) $\downarrow$ &        \bftab 0.008 $\pm$ .012 (1) &         0.056 $\pm$ .051 (3) &  0.211 $\pm$ .246 (5) &  0.084 $\pm$ .090 (4) &   0.035 $\pm$ .048 (2) \\
\bottomrule
\end{tabular}
}
\caption{\textbf{Binary Classification Performance.} We report macro F1-scores (mean $\pm$ stdev over $10$ trials) on test data with optimized hyperparameters. The rank of each method is presented in brackets. The datasets are sorted by the number of features.}

\label{tab:eval-results-binary}
\end{table*}

\begin{table*}[tb]
\centering
\small
\resizebox{0.9\textwidth}{!}{
\begin{tabular}{lccccc}
\toprule
{} &                           \multicolumn{2}{l}{Gradient-Based} &            \multicolumn{2}{l}{Non-Greedy} &                       \multicolumn{1}{l}{Greedy} \\ \cmidrule(lr){2-3}  \cmidrule(lr){4-5} \cmidrule(lr){6-6}
  
{} &                           \multicolumn{1}{l}{GradTree (ours)} &                   \multicolumn{1}{l}{DNDT} &            \multicolumn{1}{l}{GeneticTree} &                          \multicolumn{1}{l}{DL8.5 (Optimal)} &                       \multicolumn{1}{l}{CART} \\
\midrule
Iris                 &  \bftab 0.938 $\pm$ .057 (1) &         0.870 $\pm$ .063 (5) &  0.912 $\pm$ .055 (3) &  0.909 $\pm$ .046 (4) &         0.937 $\pm$ .046 (2) \\
Balance Scale        &  \bftab 0.593 $\pm$ .045 (1) &         0.475 $\pm$ .104 (5) &  0.529 $\pm$ .043 (3) &  0.525 $\pm$ .039 (4) &         0.574 $\pm$ .030 (2) \\
Car                  &         0.440 $\pm$ .085 (3) &         0.485 $\pm$ .064 (2) &  0.306 $\pm$ .068 (4) &  0.273 $\pm$ .063 (5) &  \bftab 0.489 $\pm$ .094 (1) \\
Glass                &         0.560 $\pm$ .090 (3) &         0.434 $\pm$ .072 (5) &  0.586 $\pm$ .090 (2) &  0.501 $\pm$ .100 (4) &  \bftab 0.663 $\pm$ .086 (1) \\
Contraceptive        &  \bftab 0.496 $\pm$ .050 (1) &         0.364 $\pm$ .050 (3) &  0.290 $\pm$ .048 (5) &  0.292 $\pm$ .036 (4) &         0.384 $\pm$ .075 (2) \\
Solar Flare          &         0.151 $\pm$ .033 (3) &  \bftab 0.171 $\pm$ .051 (1) &  0.146 $\pm$ .018 (4) &  0.144 $\pm$ .034 (5) &         0.157 $\pm$ .022 (2) \\
Wine                 &  \bftab 0.933 $\pm$ .031 (1) &         0.858 $\pm$ .041 (4) &  0.888 $\pm$ .039 (3) &  0.852 $\pm$ .022 (5) &         0.907 $\pm$ .042 (2) \\
Zoo                  &         0.874 $\pm$ .111 (3) &             $n>12$ &  0.782 $\pm$ .111 (4) &  0.911 $\pm$ .106 (2) &  \bftab 0.943 $\pm$ .076 (1) \\
Lymphography         &  \bftab 0.610 $\pm$ .191 (1) &             $n>12$ &  0.381 $\pm$ .124 (4) &  0.574 $\pm$ .196 (2) &         0.548 $\pm$ .154 (3) \\
Segment              &         0.941 $\pm$ .009 (2) &             $n>12$ &  0.715 $\pm$ .114 (4) &  0.808 $\pm$ .013 (3) &  \bftab 0.963 $\pm$ .010 (1) \\
Dermatology          &         0.930 $\pm$ .030 (2) &             $n>12$ &  0.785 $\pm$ .126 (4) &  0.885 $\pm$ .036 (3) &  \bftab 0.957 $\pm$ .026 (1) \\
Landsat              &         0.807 $\pm$ .011 (2) &             $n>12$ &  0.628 $\pm$ .084 (4) &  0.783 $\pm$ .008 (3) &  \bftab 0.835 $\pm$ .011 (1) \\
Annealing            &         0.638 $\pm$ .126 (3) &             $n>12$ &  0.218 $\pm$ .053 (4) &  0.787 $\pm$ .121 (2) &  \bftab 0.866 $\pm$ .094 (1) \\
Splice               &         0.873 $\pm$ .030 (2) &             $n>12$ &  0.486 $\pm$ .157 (3) &      $>$ 60 min &  \bftab 0.881 $\pm$ .021 (1) \\
\midrule
Mean Reciprocal Rank (MRR) $\uparrow$  &         0.619 $\pm$ .303 (2) &         0.383 $\pm$ .293 (3) &  0.288 $\pm$ .075 (5) &  0.315 $\pm$ .116 (4) &  \bftab 0.774 $\pm$ .274 (1) \\
Mean Relative Diff. (MRD) $\downarrow$ &         0.069 $\pm$ .102 (2) &         0.188 $\pm$ .200 (3) &  0.521 $\pm$  .749 (5) &  0.215 $\pm$ .249 (4) &  \bftab 0.040 $\pm$ .081 (1) \\


\bottomrule
\end{tabular}
}
\caption{\textbf{Multi-Class Classification Performance.} We report macro F1-scores (mean $\pm$ stdev over $10$ trials) on test data with optimized hyperparameters. The rank of each method is presented in brackets. The datasets are sorted by the number of features.}

\label{tab:eval-results-multi}
\end{table*}

\paragraph{Datasets and Preprocessing}
The experiments were conducted on several benchmark datasets, mainly from the UCI repository \citep{ml_repository}.
For all datasets, we performed a standard preprocessing: Similar to \citet{neural_oblivious}, we applied leave-one-out encoding to all categorical features and further performed a quantile transform, making each feature follow a normal distribution. 
We used a $80\%/20\%$ train-test split for all datasets. To account for class imbalance, we rebalanced datasets using SMOTE~\citep{chawla2002smote} if the minority class accounts for less than $\frac{25}{c-1} \%$ of the data, where $c$ is the number of classes.
For GradTree and DNDT, we used $20\%$ of the training data as validation data for early stopping.
As DL8.5 requires binary features, we discretized numeric features using quantile binning with $5$ bins and one-hot encoded categorical features.
Details and sources of the datasets are available in Appendix~\ref{A:dataset}

\paragraph{Methods}
We compared GradTree to the most prominent approach from each category (see Section \ref{sec:relatedwork}) to ensure a concise, yet holistic evaluation focusing on hard, axis-aligned DTs. Specifically, we selected the following methods:
\begin{itemize}
    \item \textbf{CART}: We use the sklearn~\citep{scikit-learn} implementation, which uses an optimized version of the CART algorithm. 
    CART typically employs the Gini impurity measure, but we additionally allowed entropy.
    \item \textbf{Evolutionary DTs}: We use GeneticTree~\citep{genetictree_code} for an efficient learning of DTs with a genetic algorithm. 
    \item \textbf{DNDT}: We use the official DNDT implementation~\citep{dndt_code}. For a fair comparison, we enforce binary trees by setting the number of cut points to $1$ and ensure hard splits during inference.
    As suggested by \citet{dndt}, we limited DNDTs to datasets with no more than $12$ features, due to scalability issues.
    \item \textbf{DL8.5 (Optimal DTs)}: We use the official DL8.5 implementation~\citep{pydl85_code} including improvements from MurTree~\citep{murtree} which reduces the runtime significantly.

\end{itemize}

GradTree is implemented in Python using TensorFlow\footnote{The code of our implementation is available under \url{https://github.com/s-marton/GradTree}.} 
To ensure a fair comparison, we further applied a simple post-hoc pruning for GradTree to remove all branches with zero samples based on one pass of the training data. Similar to DNDT, we used a cross-entropy loss.

\paragraph{Hyperparameters}
We conducted a random search with cross-validation to determine the optimal hyperparameters. 
The complete list of relevant hyperparameters for each approach along with additional details on the selection are in the Appendix~\ref{A:hyperparams}.

\subsection{Results}\label{ssec:experiment_results}

\paragraph{GradTree outperforms existing DT learners for binary classification}
First, we evaluated the performance of GradTree against existing approaches on the benchmark datasets in terms of the macro F1-Score, which inherently considers class imbalance. We report the relative difference to the best model (MRD) and mean reciprocal rank (MRR), following the approach of \citet{dndt}.
Overall, GradTree outperformed existing approaches for binary classification tasks (best MRR of $0.758$ and MRD of $0.008$) and achieved competitive results for multi-class tasks (second-best MRR of $0.619$ and MRD of $0.069$). 
More specifically, GradTree significantly outperformed state-of-the-art non-greedy DT methods, including DNDTs as our gradient-based benchmark.

For binary classification (Table~\ref{tab:eval-results-binary}), GradTree demonstrated superior performance over CART, achieving the best performance on $13$ datasets as compared to only $6$ datasets for CART. 
Notably, the performance difference between GradTree and existing methods was substantial for several datasets, such as \emph{Echocardiogram}, \emph{Heart Disease} and \emph{Absenteeism}.
For multi-class datasets, GradTree achieved the second-best overall performance. 
While GradTree still achieved a superior performance for low-dimensional datasets (top part of Table~\ref{tab:eval-results-multi}), CART achieved the best results for high-dimensional datasets with a high number of classes.
We can explain this by the dense representation used for the gradient-based optimization. Using our representation, the difficulty of the optimization task increases with the number of features (more parameters at each internal node) and the number of classes (more parameters at each leaf node).
In future work, we aim to optimize our dense representation, e.g., by using parameter sharing.

\paragraph{GradTree has a small effective tree size}
The effective tree size (= size after pruning) of GradTree is smaller than CART for binary and marginally higher for multi-class tasks (Table~\ref{tab:eval-results-ext-summary}). Only the tree size for GeneticTree is significantly smaller, which is caused by the complexity penalty of the genetic algorithm. DL8.5 also has a smaller average tree size than CART and GradTree. 
We can attribute this to DL8.5 being only feasible up to a depth of $4$ due to the high computational complexity.
The tree size for DNDTs scales with the number of features (and classes) which quickly results in large trees. Furthermore, pruning DNDTs is non-trivial due to the use of the Kronecker product (it is not sufficient to prune subtrees bottom-up).

\begin{table}[t]
\centering
\small
\begin{tabular}{L{1.25cm}rrR{0.6cm}R{0.6cm}rr}
\toprule 
\multirow{2}{*}{} & \multicolumn{2}{l}{Tree Size} & \multicolumn{2}{l}{\begin{tabular}[c]{@{}l@{}}Train-Test\\ Difference\end{tabular}}& \multicolumn{2}{L{1.925cm}}{Default Setting (MRR $\uparrow$)}\\ \cmidrule(lr){2-3}  \cmidrule(lr){4-5} \cmidrule(lr){6-7} 
& \multicolumn{1}{L{0.6cm}}{Binary} & \multicolumn{1}{L{0.6cm}}{Multi} & \multicolumn{1}{L{0.65cm}}{Binary} & \multicolumn{1}{L{0.65cm}}{Multi} & \multicolumn{1}{l}{Binary} & \multicolumn{1}{l}{Multi} \\

\midrule
GradTree              & \phantom{0}54           & \phantom{0}86   &  0.051     & 0.174    & \bftab 0.670     & 0.470 \\ 
DNDT                    & 887                     & 907             & 0.039      & 0.239    & 0.306     & 0.295 \\
GeneticTree             & \phantom{00}7           & \phantom{0}20   & 0.204      & 0.258    & 0.427     & 0.315 \\
DL8.5                   & \phantom{0}28           & \phantom{0}29   & 0.202      & 0.260    & 0.371     & 0.331 \\
CART                    & \phantom{0}67           & \phantom{0}76   & 0.183      & 0.247    & 0.571     & \bftab  0.929 \\
\bottomrule
\end{tabular}

\caption{\textbf{Summarized Results.} Left: Average tree size. Mid: Mean difference between train and test performance as overfitting indicator. Right: Test performance with default parameters. Detailed results are in Table~\ref{tab:eval-results_noHPO}-\ref{tab:eval-results-tree-size}.}
\label{tab:eval-results-ext-summary}
\end{table}

\paragraph{GradTree is robust to overfitting}
We can observe that gradient-based approaches were more robust and less prone to overfitting compared to a greedy optimization with CART and alternative non-greedy methods. We measure overfitting by the difference between the mean train and test performance (see Table~\ref{tab:eval-results-ext-summary}). 
For binary tasks, GradTree exhibits a train-test performance difference of $0.051$, considerably smaller than that of CART ($0.183$), GeneticTree ($0.204$), and DL8.5 ($0.202$).
DNDTs, which are also gradient-based, achieved an even smaller difference of $0.039$. For multi-class tasks, the difference was significantly smaller for GradTree compared to any other approach.

\paragraph{GradTree does not rely on extensive hyperparameter optimization}
Besides their interpretability, a distinct advantage of DTs over more sophisticated models is that they typically do not rely on an extensive hyperparameter optimization.
In this experiment, we show that the same is true for GradTree by evaluating the performance with default configurations (see Table~\ref{tab:eval-results-ext-summary}). When using the default parameters, GradTree still outperformed the other methods on binary tasks (highest MRR and most wins) and achieved competitive results for multi-class tasks (second-highest MRR).

\paragraph{GradTree is efficient for large and high-dimensional datasets}
For each dataset, a greedy optimization using CART was substantially faster than other methods, taking less than a second.
Nevertheless, for most datasets, training GradTree took less than $30$ seconds (mean runtime of $35$ seconds).
DNDT had comparable runtimes to GradTree. 
For most datasets, DL8.5 had a low runtime of less than 10 seconds.
However, scalability issues become apparent with DL8.5, especially with an increasing number of features and samples. Its runtime surpassed GradTree on various datasets, taking around 300 seconds for \emph{Credit Card} and exceeding 830 seconds for \emph{Landsat}.
For \emph{Splice}, DL8.5 did not find a solution within $60$ minutes and the optimization was terminated. Detailed runtimes are in Table~\ref{tab:eval-results-runtime}.

\paragraph{ST entmax outperforms alternative methods}\label{sssec:ablation}
In an ablation study (Table~\ref{tab:ablation_study}), we evaluated our design choice of utilizing an ST operator directly after an entmax transformation to address the non-differentiability of DTs. We contrasted this against alternative strategies found in the literature.
Our approach notably surpassed ST Gumbel Softmax~\citep{gumbel} and outperformed the temperature annealing technique proposed by \citet{node_gam} to gradually turn the entmax one-hot.

\begin{table}[t]
\centering
\small
\begin{tabular}{llrrr}
\toprule 

 & & \multicolumn{1}{L{1.45cm}}{ST Entmax (ours)} & \multicolumn{1}{L{1.15cm}}{ST Gumbel} & \multicolumn{1}{L{1.35cm}}{Temp. Annealing} \\

\midrule

\multirow{2}{*}{Default} & Binary         & \bftab 0.764      & 0.560     & 0.757 \\ 
                        & Multi    & \bftab 0.638      & 0.272    & 0.602 \\ \cmidrule(lr){1-2}
                             
\multirow{2}{*}{Optimized}    & Binary         & \bftab 0.771      & 0.569     & 0.759 \\                 
                        & Multi    & \bftab 0.699      & 0.297     & 0.601 \\                   
\bottomrule
\end{tabular}


\caption{\textbf{Ablation Study.} We compare our approach to deal with the non-differentiable nature of DTs with alternative methods, reporting the average macro F1-scores over $10$ trials with optimized and default hyperparameters. The complete results are listed in Table~\ref{tab:ablation_HPO}-\ref{tab:ablation_noHPO}.} 
\label{tab:ablation_study}
\end{table}

\section{Conclusion and Future Work}\label{sec:conclusion}

In this paper, we proposed GradTree, a novel method for learning hard, axis-aligned DTs based on a joint optimization of all tree parameters with gradient descent. 
Our empirical evaluations indicate that GradTree excels over existing methods in binary tasks and offers competitive performance in multi-class tasks.
The substantial performance increase achieved by GradTree across multiple datasets highlights its importance as a noteworthy contribution to the existing repertoire of DT learning methods.

Moreover, gradient-based optimization provides greater flexibility, allowing for easy integration of custom loss functions tailored to specific application scenarios.
Another advantage is the ability to relearn the threshold value as well as the split index. Therefore, GradTree is suitable for dynamic environments, such as online learning tasks.

Currently, GradTree employs conventional post-hoc pruning. 
In future work, we want to consider pruning already during the training, for instance through a learnable choice parameter to decide if a node is pruned, similar to \citet{zantedeschi2021learning}.
Although our focus was on stand-alone DTs aiming for intrinsic interpretability, GradTree holds potential as a foundational method for learning hard, axis-aligned tree ensembles end-to-end via gradient descent. Exploring this performance-interpretability trade-off is an interesting direction for future research.

\bibliography{gradtree}

\onecolumn
\newpage

\appendix

\section{Gradient Descent Optimization}
We use stochastic gradient descent (SGD) to minimize the loss function of GradTree, which is outlined in Algorithm~\ref{alg:train}. We use backpropagation to calculate the gradients in Line 11-13.
Furthermore, our implementation optimizes Algorithm~\ref{alg:train} by exploiting common SGD techniques, including mini-batch calculation and momentum using the Adam optimizer~\citep{kingma2014adam}. We further apply weight averaging~\citep{izmailov2018averaging} over 5 consecutive checkpoints, similar to \citet{neural_oblivious}. Our novel tree pass function allows formulating Line 7-9 as a single set of matrix operations for an entire batch, which results in a very efficient optimization.
Moreover, we implement an early stopping procedure based on the validation loss. To avoid bad initial parametrizations during the initialization, we additionally implement random restarts where the best parameters are selected based on the validation loss.

\begin{algorithm}[h]
\caption{Gradient Descent Training for Decision Trees}   \label{alg:train} 

\begin{algorithmic}[1] 
    \Function{trainDT}{$I, T, L, X, \boldsymbol{y}, n, c, d, \xi$}
        \State{$I \sim \mathcal{U}\left(-\sqrt{\frac{6}{2^{2d-1} + n}},\sqrt{\frac{6}{2^{2d-1} + n}}\right)$} 
        \State{$T \sim \mathcal{U}\left(-\sqrt{\frac{6}{2^{2d-1} + n}},\sqrt{\frac{6}{2^{2d-1} + n}}\right)$} 
        \State{$L \sim \mathcal{U}\left(-\sqrt{\frac{6}{2^{2d} + c}},\sqrt{\frac{6}{2^{2d} + c}}\right)$} 
        
        \item[]

        \For{$i = 1,\dots,\xi$}
            \State{$\boldsymbol{\hat{y}} \leftarrow \emptyset$} 
            \For{$j = 1,\dots,|X|$}
                \State{$\hat{y}_j = $ \Call{pass}{$I, T, L, X_j$}}
                \State{$\boldsymbol{\hat{y}} \leftarrow \boldsymbol{\hat{y}} \cup \hat{y}_j$}
            \EndFor
            \State{$I \leftarrow I + \eta \frac{\partial}{\partial I} \mathcal{L}(\boldsymbol{y}, \boldsymbol{\hat{y}})$} \Comment{Calculate gradients with backpropagation} 
            \State{$T \leftarrow T + \eta \frac{\partial}{\partial T} \mathcal{L}(\boldsymbol{y}, \boldsymbol{\hat{y}})$}\Comment{Calculate gradients with backpropagation} 
            \State{$L \leftarrow L + \eta \frac{\partial}{\partial L} \mathcal{L}(\boldsymbol{y}, \boldsymbol{\hat{y}})$}\Comment{Calculate gradients with backpropagation} 

        \EndFor

    \EndFunction
\end{algorithmic}

\end{algorithm}

\section{Additional Results}

\subsection{Ablation Study}

\begin{table}[H]

\centering
\begin{tabular}{lccc}
    \toprule
    & ST Entmax (ours) & ST Gumbel Softmax & Temperature Annealing \\
    \midrule
        Blood Transfusion & \bftab 0.628 $\pm$ .036 (1) & 0.482 $\pm$ .083 (3) & 0.606 $\pm$ .073 (2) \\
        Banknote Authentication & \bftab 0.987 $\pm$ .007 (1) & 0.770 $\pm$ .144 (3) & 0.946 $\pm$ .041 (2) \\
        Titanic & \bftab 0.776 $\pm$ .025 (1) & 0.543 $\pm$ .093 (3) & 0.762 $\pm$ .036 (2) \\
        Raisins & 0.840 $\pm$ .022 (2) & 0.792 $\pm$ .069 (3) & \bftab 0.846 $\pm$ .028 (1) \\
        Rice & 0.926 $\pm$ .007 (2) & 0.859 $\pm$ .058 (3) & \bftab 0.927 $\pm$ .006 (1) \\
        Echocardiogram & \bftab 0.658 $\pm$ .113 (1) & 0.574 $\pm$ .108 (3) & 0.619 $\pm$ .151 (2) \\
        Wisconsin Diagnostic Breast Cancer & \bftab 0.904 $\pm$ .022 (1) & 0.844 $\pm$ .059 (3) & 0.893 $\pm$ .031 (2) \\
        Loan House & \bftab 0.714 $\pm$ .041 (1) & 0.522 $\pm$ .098 (3) & 0.692 $\pm$ .051 (2) \\
        Heart Failure & \bftab 0.750 $\pm$ .070 (1) & 0.556 $\pm$ .095 (3) & 0.749 $\pm$ .060 (2) \\
        Heart Disease & \bftab 0.779 $\pm$ .047 (1) & 0.607 $\pm$ .136 (3) & 0.754 $\pm$ .035 (2) \\
        Adult & \bftab 0.743 $\pm$ .034 (1) & 0.583 $\pm$ .029 (3) & 0.737 $\pm$ .045 (2) \\
        Bank Marketing & \bftab 0.640 $\pm$ .027 (1) & 0.370 $\pm$ .052 (3) & 0.611 $\pm$ .053 (2) \\
        Congressional Voting & \bftab 0.950 $\pm$ .021 (1) & 0.710 $\pm$ .201 (3) & 0.946 $\pm$ .026 (2) \\
        Absenteeism & \bftab 0.626 $\pm$ .047 (1) & 0.489 $\pm$ .059 (3) & 0.604 $\pm$ .036 (2) \\
        Hepatitis & \bftab 0.608 $\pm$ .078 (1) & 0.395 $\pm$ .138 (3) & 0.576 $\pm$ .122 (2) \\
        German & 0.592 $\pm$ .068 (2) & 0.486 $\pm$ .072 (3) & \bftab 0.634 $\pm$ .028 (1) \\
        Mushroom & \bftab 1.000 $\pm$ .001 (1) & 0.682 $\pm$ .063 (3) & 0.996 $\pm$ .005 (2) \\
        Credit Card & \bftab 0.674 $\pm$ .014 (1) & 0.488 $\pm$ .019 (3) & 0.668 $\pm$ .015 (2) \\
        Horse Colic & 0.842 $\pm$ .039 (2) & 0.442 $\pm$ .126 (3) & \bftab 0.843 $\pm$ .039 (1) \\
        Thyroid & \bftab 0.905 $\pm$ .010 (1) & 0.449 $\pm$ .087 (3) & 0.888 $\pm$ .017 (2) \\
        Cervical Cancer & 0.521 $\pm$ .043 (2) & 0.333 $\pm$ .191 (3) & \bftab 0.523 $\pm$ .042 (1) \\
        Spambase & \bftab 0.903 $\pm$ .025 (1) & 0.541 $\pm$ .093 (3) & 0.875 $\pm$ .019 (2) \\
        \midrule
        Mean $\uparrow$ & \bftab 0.771 $\pm$ .036 (1) & 0.569 $\pm$ .094 (3) & 0.759 $\pm$ .044 (2) \\    
        Mean Reciprocal Rank (MRR) $\uparrow$  & \bftab 0.886 $\pm$ .214 (1) & 0.333 $\pm$ .000 (3) & 0.613 $\pm$ .214 (2) \\    
    \midrule    
    \midrule
        Iris & \bftab 0.938 $\pm$ .057 (1) & 0.743 $\pm$ .224 (3) & 0.930 $\pm$ .048 (2) \\
        Balance Scale & \bftab 0.593 $\pm$ .045 (1) & 0.372 $\pm$ .073 (3) & 0.561 $\pm$ .050 (2) \\
        Car & \bftab 0.440 $\pm$ .085 (1) & 0.226 $\pm$ .067 (3) & 0.334 $\pm$ .070 (2) \\
        Glass & \bftab 0.560 $\pm$ .090 (1) & 0.209 $\pm$ .100 (3) & 0.467 $\pm$ .077 (2) \\
        Contraceptive & \bftab 0.496 $\pm$ .050 (1) & 0.324 $\pm$ .064 (3) & 0.422 $\pm$ .074 (2) \\
        Solar Flare & \bftab 0.151 $\pm$ .033 (1) & 0.065 $\pm$ .042 (3) & 0.150 $\pm$ .021 (2) \\
        Wine & \bftab 0.933 $\pm$ .031 (1) & 0.565 $\pm$ .195 (3) & 0.881 $\pm$ .043 (2) \\
        Zoo & \bftab 0.874 $\pm$ .111 (1) & 0.183 $\pm$ .093 (3) & 0.678 $\pm$ .111 (2) \\
        Lymphography & \bftab 0.610 $\pm$ .191 (1) & 0.307 $\pm$ .104 (2) & 0.303 $\pm$ .121 (3) \\
        Segment & \bftab 0.941 $\pm$ .009 (1) & 0.245 $\pm$ .082 (3) & 0.835 $\pm$ .040 (2) \\
        Dermatology & \bftab 0.930 $\pm$ .030 (1) & 0.148 $\pm$ .091 (3) & 0.757 $\pm$ .144 (2) \\
        Landsat & \bftab 0.807 $\pm$ .011 (1) & 0.349 $\pm$ .051 (3) & 0.789 $\pm$ .009 (2) \\
        Annealing & \bftab 0.638 $\pm$ .126 (1) & 0.089 $\pm$ .078 (3) & 0.597 $\pm$ .074 (2) \\
        Splice & \bftab 0.873 $\pm$ .030 (1) & 0.337 $\pm$ .033 (3) & 0.711 $\pm$ .054 (2) \\
        \midrule
        Mean $\uparrow$ & \bftab 0.699 $\pm$ .064 (1) & 0.297 $\pm$ .093 (3) & 0.601 $\pm$ .067 (2) \\
        Mean Reciprocal Rank (MRR) $\uparrow$  & \bftab 1.000 $\pm$ .000 (1) & 0.345 $\pm$ .043 (3) & 0.488 $\pm$ .043 (2) \\    
    \bottomrule
\end{tabular}
\caption{Ablation Study Optimized Parameters. We report macro F1-scores (mean $\pm$ stdev over $10$ trials) with optimized parameters. We also report the rank of each approach in brackets. The datasets are sorted by the number of features. The top part comprises binary classification tasks and the bottom part multi-class datasets.}
\label{tab:ablation_HPO}
\end{table}

\begin{table}[H]
\centering
\begin{tabular}{lccc}
    \toprule
    & ST Entmax (ours) & ST Gumbel Softmax & Temperature Annealing \\
    \midrule
        Blood Transfusion & \bftab 0.627 $\pm$ .056 (1) & 0.497 $\pm$ .086 (3) & 0.618 $\pm$ .045 (2) \\
        Banknote Authentication & \bftab 0.980 $\pm$ .008 (1) & 0.722 $\pm$ .126 (3) & 0.970 $\pm$ .007 (2) \\
        Titanic & \bftab 0.782 $\pm$ .035 (1) & 0.623 $\pm$ .096 (3) & 0.769 $\pm$ .033 (2) \\
        Raisins & 0.850 $\pm$ .025 (2) & 0.802 $\pm$ .055 (3) & \bftab 0.853 $\pm$ .016 (1) \\
        Rice & \bftab 0.926 $\pm$ .006 (1) & 0.802 $\pm$ .101 (3) & \bftab 0.926 $\pm$ .006 (1) \\
        Echocardiogram &\bftab  0.648 $\pm$ .130 (1) & 0.571 $\pm$ .093 (3) & 0.595 $\pm$ .121 (2) \\
        Wisconsin Diagnostic Breast Cancer & 0.902 $\pm$ .029 (2) & 0.841 $\pm$ .129 (3) & \bftab 0.903 $\pm$ .028 (1) \\
        Loan House & 0.695 $\pm$ .035 (2) & 0.490 $\pm$ .113 (3) & \bftab 0.706 $\pm$ .040 (1) \\
        Heart Failure & 0.745 $\pm$ .063 (2) & 0.489 $\pm$ .095 (3) & \bftab 0.761 $\pm$ .055 (1) \\
        Heart Disease & \bftab 0.736 $\pm$ .069 (1) & 0.473 $\pm$ .148 (3) & 0.721 $\pm$ .049 (2) \\
        Adult & \bftab 0.749 $\pm$ .028 (1) & 0.601 $\pm$ .034 (3) & 0.737 $\pm$ .042 (2) \\
        Bank Marketing & \bftab 0.626 $\pm$ .017 (1) & 0.389 $\pm$ .084 (3) & \bftab 0.626 $\pm$ .010 (1) \\
        Congressional Voting & \bftab 0.953 $\pm$ .021 (1) & 0.642 $\pm$ .161 (3) & \bftab 0.953 $\pm$ .021 (2) \\
        Absenteeism & 0.620 $\pm$ .050 (2) & 0.452 $\pm$ .056 (3) & \bftab 0.625 $\pm$ .065 (1) \\
        Hepatitis & \bftab 0.609 $\pm$ .103 (1) & 0.390 $\pm$ .125 (3) & 0.574 $\pm$ .124 (2) \\
        German & 0.584 $\pm$ .057 (2) & 0.511 $\pm$ .086 (3) & \bftab 0.596 $\pm$ .052 (1) \\
        Mushroom & \bftab 0.997 $\pm$ .003 (1) & 0.701 $\pm$ .083 (3) & 0.949 $\pm$ .095 (2) \\
        Credit Card & \bftab 0.676 $\pm$ .009 (1) & 0.479 $\pm$ .020 (3) & 0.669 $\pm$ .005 (2) \\
        Horse Colic & 0.842 $\pm$ .033 (2) & 0.458 $\pm$ .101 (3) &\bftab  0.857 $\pm$ .050 (1) \\
        Thyroid & \bftab 0.872 $\pm$ .015 (1) & 0.457 $\pm$ .099 (3) & 0.866 $\pm$ .013 (2) \\
        Cervical Cancer & \bftab 0.496 $\pm$ .047 (1) & 0.354 $\pm$ .160 (3) & 0.479 $\pm$ .048 (2) \\
        Spambase & \bftab 0.893 $\pm$ .015 (1) & 0.580 $\pm$ .077 (3) & 0.891 $\pm$ .008 (2) \\
    \midrule
        Mean $\uparrow$ & \bftab 0.764 $\pm$ .039 (1) & 0.560 $\pm$ .097 (3) & 0.757 $\pm$ .042 (2) \\
        Mean Reciprocal Rank (MRR) $\uparrow$ & \bftab 0.841 $\pm$ .238 (1) & 0.333 $\pm$ .000 (3) & 0.705 $\pm$ .252 (2) \\    
    \midrule
    \midrule
    Iris & \bftab 0.922 $\pm$ .068 (1) & 0.745 $\pm$ .207 (3) & 0.916 $\pm$ .053 (2) \\
    Balance Scale & \bftab 0.565 $\pm$ .043 (1) & 0.323 $\pm$ .075 (3) & 0.547 $\pm$ .037 (2) \\
    Car & \bftab 0.408 $\pm$ .066 (1) & 0.249 $\pm$ .039 (3) & 0.387 $\pm$ .095 (2) \\
    Glass & \bftab 0.482 $\pm$ .110 (1) & 0.139 $\pm$ .092 (3) & 0.405 $\pm$ .110 (2) \\
    Contraceptive & 0.398 $\pm$ .054 (2) & 0.320 $\pm$ .048 (3) & \bftab 0.407 $\pm$ .056 (1) \\
    Solar Flare & \bftab 0.152 $\pm$ .031 (1) & 0.062 $\pm$ .051 (3) & 0.140 $\pm$ .049 (2) \\
    Wine & 0.861 $\pm$ .057 (2) & 0.451 $\pm$ .104 (3) & \bftab 0.884 $\pm$ .040 (1) \\
    Zoo & \bftab 0.839 $\pm$ .136 (1) & 0.144 $\pm$ .119 (3) & 0.709 $\pm$ .151 (2) \\
    Lymphography & \bftab 0.463 $\pm$ .162 (1) & 0.291 $\pm$ .133 (3) & 0.337 $\pm$ .156 (2) \\
    Segment & \bftab 0.918 $\pm$ .014 (1) & 0.177 $\pm$ .116 (3) & 0.906 $\pm$ .026 (2) \\
    Dermatology & \bftab 0.903 $\pm$ .025 (1) & 0.211 $\pm$ .137 (3) & 0.839 $\pm$ .085 (2) \\
    Landsat & \bftab 0.771 $\pm$ .019 (1) & 0.343 $\pm$ .066 (3) & 0.769 $\pm$ .027 (2) \\
    Annealing & \bftab 0.496 $\pm$ .111 (1) & 0.073 $\pm$ .049 (3) & 0.478 $\pm$ .078 (2) \\
    Splice & \bftab 0.748 $\pm$ .168 (1) & 0.283 $\pm$ .046 (3) & 0.703 $\pm$ .056 (2) \\
    \midrule
    Mean $\uparrow$ & \bftab 0.638 $\pm$ .076 (1) & 0.272 $\pm$ .092 (3) & 0.602 $\pm$ .073 (2) \\
    Mean Reciprocal Rank (MRR) $\uparrow$ & \bftab 0.929 $\pm$ .175 (1) & 0.333 $\pm$ .000 (3) & 0.571 $\pm$ .175 (2) \\

    \bottomrule
\end{tabular}
\caption{Ablation Study Default Parameters. We report macro F1-scores (mean $\pm$ stdev over $10$ trials) with default parameters. We also report the rank of each approach in brackets. The datasets are sorted by the number of features. The top part comprises binary classification tasks and the bottom part multi-class datasets.}
\label{tab:ablation_noHPO}
\end{table}

\subsection{Default Parameter Performance}

\begin{table}[H]

\centering
\resizebox{1.0\columnwidth}{!}{
\begin{tabular}{lccccc}
\toprule
{} &                           \multicolumn{2}{l}{Gradient-Based} &            \multicolumn{2}{l}{Non-Greedy} &                       \multicolumn{1}{l}{Greedy} \\ \cmidrule(lr){2-3}  \cmidrule(lr){4-5} \cmidrule(lr){6-6}
  
{} &                           \multicolumn{1}{l}{GradTree (ours)} &                   \multicolumn{1}{l}{DNDT} &            \multicolumn{1}{l}{GeneticTree} &                          \multicolumn{1}{l}{DL8.5 (Optimal)} &                       \multicolumn{1}{l}{CART} \\
\midrule
Blood Transfusion                  &  \bftab 0.627 $\pm$ .056 (1) &         0.558 $\pm$ .059 (5) &         0.574 $\pm$ .093 (3) &         0.590 $\pm$ .034 (2) &         0.573 $\pm$ .036 (4) \\
Banknote Authentication            &         0.980 $\pm$ .008 (2) &         0.886 $\pm$ .024 (5) &         0.928 $\pm$ .022 (4) &         0.962 $\pm$ .011 (3) &  \bftab 0.981 $\pm$ .006 (1) \\
Titanic                            &  \bftab 0.782 $\pm$ .035 (1) &         0.740 $\pm$ .026 (4) &         0.763 $\pm$ .041 (2) &         0.754 $\pm$ .031 (3) &         0.735 $\pm$ .041 (5) \\
Raisins                            &         0.850 $\pm$ .025 (2) &         0.832 $\pm$ .026 (4) &  \bftab 0.859 $\pm$ .022 (1) &         0.849 $\pm$ .027 (3) &         0.811 $\pm$ .028 (5) \\
Rice                               &         0.926 $\pm$ .006 (2) &         0.902 $\pm$ .018 (5) &  \bftab 0.927 $\pm$ .005 (1) &         0.925 $\pm$ .008 (3) &         0.905 $\pm$ .010 (4) \\
Echocardiogram                     &  \bftab 0.648 $\pm$ .130 (1) &         0.543 $\pm$ .117 (5) &         0.563 $\pm$ .099 (3) &         0.609 $\pm$ .112 (2) &         0.559 $\pm$ .075 (4) \\
Wisconsin Breast Cancer &         0.902 $\pm$ .029 (3) &  \bftab 0.907 $\pm$ .037 (1) &         0.888 $\pm$ .021 (5) &         0.896 $\pm$ .021 (4) &         0.904 $\pm$ .025 (2) \\
Loan House                         &  \bftab 0.695 $\pm$ .035 (1) &         0.475 $\pm$ .043 (4) &         0.461 $\pm$ .093 (5) &         0.607 $\pm$ .045 (3) &         0.671 $\pm$ .056 (2) \\
Heart Failure                      &         0.745 $\pm$ .063 (2) &         0.580 $\pm$ .077 (5) &         0.740 $\pm$ .055 (3) &         0.692 $\pm$ .062 (4) &  \bftab 0.755 $\pm$ .060 (1) \\
Heart Disease                      &  \bftab 0.736 $\pm$ .069 (1) &                        $n>12$ &         0.704 $\pm$ .059 (3) &         0.722 $\pm$ .065 (2) &         0.670 $\pm$ .090 (4) \\
Adult                              &         0.749 $\pm$ .028 (2) &                        $n>12$ &         0.478 $\pm$ .068 (4) &         0.723 $\pm$ .011 (3) &  \bftab 0.773 $\pm$ .008 (1) \\
Bank Marketing                     &  \bftab 0.626 $\pm$ .017 (1) &                        $n>12$ &         0.473 $\pm$ .002 (4) &         0.502 $\pm$ .011 (3) &         0.616 $\pm$ .007 (2) \\
Congressional Voting               &  \bftab 0.953 $\pm$ .021 (1) &                        $n>12$ &         0.932 $\pm$ .034 (3) &         0.924 $\pm$ .043 (4) &         0.933 $\pm$ .032 (2) \\
Absenteeism                        &  \bftab 0.620 $\pm$ .050 (1) &                        $n>12$ &         0.417 $\pm$ .035 (4) &         0.587 $\pm$ .047 (2) &         0.580 $\pm$ .045 (3) \\
Hepatitis                          &         0.609 $\pm$ .103 (2) &                        $n>12$ &         0.486 $\pm$ .074 (4) &         0.586 $\pm$ .083 (3) &  \bftab 0.610 $\pm$ .123 (1) \\
German                             &         0.584 $\pm$ .057 (2) &                        $n>12$ &         0.412 $\pm$ .005 (4) &         0.556 $\pm$ .035 (3) &  \bftab 0.595 $\pm$ .028 (1) \\
Mushroom                           &         0.997 $\pm$ .003 (3) &                        $n>12$ &         0.984 $\pm$ .003 (4) &  \bftab 0.999 $\pm$ .001 (1) &         0.999 $\pm$ .001 (2) \\
Credit Card                        &         0.676 $\pm$ .009 (4) &                        $n>12$ &  \bftab 0.685 $\pm$ .004 (1) &         0.679 $\pm$ .007 (3) &         0.679 $\pm$ .007 (2) \\
Horse Colic                        &  \bftab 0.842 $\pm$ .033 (1) &                        $n>12$ &         0.794 $\pm$ .042 (2) &         0.708 $\pm$ .038 (4) &         0.758 $\pm$ .053 (3) \\
Thyroid                            &         0.872 $\pm$ .015 (2) &                        $n>12$ &         0.476 $\pm$ .101 (4) &         0.682 $\pm$ .018 (3) &  \bftab 0.912 $\pm$ .013 (1) \\
Cervical Cancer                    &         0.496 $\pm$ .047 (3) &                        $n>12$ &  \bftab 0.514 $\pm$ .034 (1) &         0.488 $\pm$ .027 (4) &         0.505 $\pm$ .033 (2) \\
Spambase                           &         0.893 $\pm$ .015 (2) &                        $n>12$ &         0.864 $\pm$ .014 (3) &         0.863 $\pm$ .011 (4) &  \bftab 0.917 $\pm$ .011 (1) \\
\midrule
Mean $\uparrow$                            &  \bftab 0.764 $\pm$ .144 (1) &                             - &         0.678 $\pm$ .197 (4) &         0.723 $\pm$ .154 (3) &         0.747 $\pm$ .153 (2) \\
Mean Reciprocal Rank (MRR) $\uparrow$               &  \bftab 0.670 $\pm$ .289 (1) &         0.306 $\pm$ .262 (5) &         0.427 $\pm$ .287 (3) &         0.371 $\pm$ .164 (4) &         0.571 $\pm$ .318 (2) \\
\midrule \midrule
Iris                 &  \bftab 0.938 $\pm$ .039 (1) &  0.870 $\pm$ .051 (5) &  0.912 $\pm$ .038 (3) &  0.909 $\pm$ .046 (4) &         0.937 $\pm$ .036 (2) \\
Balance Scale        &         0.575 $\pm$ .031 (2) &  0.542 $\pm$ .062 (3) &  0.539 $\pm$ .026 (4) &  0.525 $\pm$ .039 (5) &  \bftab 0.581 $\pm$ .025 (1) \\
Car                  &         0.389 $\pm$ .062 (3) &  0.426 $\pm$ .065 (2) &  0.321 $\pm$ .069 (4) &  0.273 $\pm$ .063 (5) &  \bftab 0.571 $\pm$ .133 (1) \\
Glass                &         0.484 $\pm$ .099 (5) &  0.498 $\pm$ .096 (4) &  0.566 $\pm$ .152 (2) &  0.501 $\pm$ .100 (3) &  \bftab 0.668 $\pm$ .085 (1) \\
Contraceptive        &  \bftab 0.472 $\pm$ .039 (1) &  0.324 $\pm$ .045 (3) &  0.311 $\pm$ .053 (4) &  0.292 $\pm$ .036 (5) &         0.450 $\pm$ .064 (2) \\
Solar Flare          &         0.145 $\pm$ .017 (3) &  0.142 $\pm$ .025 (5) &  0.150 $\pm$ .026 (2) &  0.144 $\pm$ .034 (4) &  \bftab 0.180 $\pm$ .026 (1) \\
Wine                 &         0.895 $\pm$ .035 (3) &  0.853 $\pm$ .042 (4) &  0.898 $\pm$ .041 (2) &  0.852 $\pm$ .022 (5) &  \bftab 0.907 $\pm$ .042 (1) \\
Zoo                  &         0.827 $\pm$ .162 (3) &                 $n>12$ &  0.802 $\pm$ .094 (4) &  0.911 $\pm$ .106 (2) &  \bftab 0.943 $\pm$ .076 (1) \\
Lymphography         &         0.472 $\pm$ .192 (3) &                 $n>12$ &  0.397 $\pm$ .211 (4) &  0.574 $\pm$ .196 (2) &  \bftab 0.606 $\pm$ .176 (1) \\
Segment              &         0.921 $\pm$ .014 (2) &                 $n>12$ &  0.736 $\pm$ .064 (4) &  0.808 $\pm$ .013 (3) &  \bftab 0.950 $\pm$ .006 (1) \\
Dermatology          &         0.876 $\pm$ .041 (3) &                 $n>12$ &  0.818 $\pm$ .161 (4) &  0.885 $\pm$ .036 (2) &  \bftab 0.951 $\pm$ .024 (1) \\
Landsat              &         0.791 $\pm$ .012 (2) &                 $n>12$ &  0.647 $\pm$ .054 (4) &  0.783 $\pm$ .008 (3) &  \bftab 0.835 $\pm$ .011 (1) \\
Annealing            &         0.632 $\pm$ .155 (3) &                 $n>12$ &  0.198 $\pm$ .050 (4) &  0.787 $\pm$ .121 (2) &  \bftab 0.873 $\pm$ .092 (1) \\
Splice               &         0.869 $\pm$ .017 (2) &                 $n>12$ &  0.611 $\pm$ .095 (3) &               $>$ 60 min &  \bftab 0.878 $\pm$ .028 (1) \\
\midrule
Mean $\uparrow$              &         0.638 $\pm$ .244 (2) &  -                     &  0.565 $\pm$ .256 (4) &  0.634 $\pm$ .268 (3) &  \bftab 0.738 $\pm$ .235 (1) \\
Mean Reciprocal Rank (MRR) $\uparrow$ &         0.470 $\pm$ .239 (2) &  0.295 $\pm$ .106 (5) &  0.315 $\pm$ .104 (4) &  0.331 $\pm$ .128 (3) &  \bftab 0.929 $\pm$ .182 (1) \\
\bottomrule
\end{tabular}
}
\caption{Performance Comparison Default Hyperparameters. We report macro F1-scores (mean $\pm$ stdev over $10$ trials) with default parameters. We also report the rank of each approach in brackets. The datasets are sorted by the number of features. The top part comprises binary classification tasks and the bottom part multi-class datasets.}
\label{tab:eval-results_noHPO}
\end{table}

\subsection{Train Data Performance}

\begin{table}[H]

\centering
\resizebox{1.0\columnwidth}{!}{
\begin{tabular}{lccccc}
\toprule
{} &                           \multicolumn{2}{l}{Gradient-Based} &            \multicolumn{2}{l}{Non-Greedy} &                       \multicolumn{1}{l}{Greedy} \\ \cmidrule(lr){2-3}  \cmidrule(lr){4-5} \cmidrule(lr){6-6}
  
{} &                           \multicolumn{1}{l}{GradTree (ours)} &                   \multicolumn{1}{l}{DNDT} &            \multicolumn{1}{l}{GeneticTree} &                          \multicolumn{1}{l}{DL8.5 (Optimal)} &                       \multicolumn{1}{l}{CART} \\
\midrule
Blood Transfusion                  &         0.686 $\pm$ .032 (3) &  0.552 $\pm$ .085 (5) &  0.615 $\pm$ .122 (4) &         0.711 $\pm$ .045 (2) &  \bftab 0.832 $\pm$ .040 (1) \\
Banknote Authentication            &         0.995 $\pm$ .003 (2) &  0.907 $\pm$ .012 (5) &  0.933 $\pm$ .016 (4) &         0.967 $\pm$ .003 (3) &  \bftab 0.999 $\pm$ .001 (1) \\
Titanic                            &         0.761 $\pm$ .128 (5) &  0.791 $\pm$ .025 (4) &  0.908 $\pm$ .023 (3) &  \bftab 0.970 $\pm$ .006 (1) &         0.949 $\pm$ .012 (2) \\
Raisins                            &  \bftab 0.892 $\pm$ .024 (1) &  0.829 $\pm$ .027 (5) &  0.863 $\pm$ .004 (4) &         0.889 $\pm$ .004 (2) &         0.866 $\pm$ .004 (3) \\
Rice                               &         0.925 $\pm$ .003 (3) &  0.913 $\pm$ .012 (5) &  0.920 $\pm$ .002 (4) &  \bftab 0.930 $\pm$ .002 (1) &         0.927 $\pm$ .002 (2) \\
Echocardiogram                     &         0.827 $\pm$ .088 (4) &  0.817 $\pm$ .034 (5) &  0.844 $\pm$ .032 (3) &         0.935 $\pm$ .013 (2) &  \bftab 0.981 $\pm$ .009 (1) \\
Wisconsin Breast Cancer &         0.947 $\pm$ .015 (2) &  0.936 $\pm$ .015 (3) &  0.907 $\pm$ .009 (5) &  \bftab 0.964 $\pm$ .005 (1) &         0.915 $\pm$ .006 (4) \\
Loan House                         &         0.735 $\pm$ .009 (4) &  0.712 $\pm$ .010 (5) &  0.897 $\pm$ .012 (3) &  \bftab 0.969 $\pm$ .006 (1) &         0.938 $\pm$ .009 (2) \\
Heart Failure                      &         0.785 $\pm$ .038 (4) &  0.773 $\pm$ .053 (5) &  0.795 $\pm$ .024 (3) &  \bftab 0.912 $\pm$ .008 (1) &         0.818 $\pm$ .021 (2) \\
Heart Disease                      &         0.851 $\pm$ .027 (4) &      $n>12$ &  0.902 $\pm$ .020 (3) &  \bftab 0.990 $\pm$ .005 (1) &         0.964 $\pm$ .010 (2) \\
Adult                              &         0.875 $\pm$ .050 (4) &      $n>12$ &  0.932 $\pm$ .013 (3) &         0.967 $\pm$ .001 (2) &  \bftab 0.973 $\pm$ .001 (1) \\
Bank Marketing                     &         0.603 $\pm$ .029 (4) &      $n>12$ &  0.971 $\pm$ .003 (3) &         0.981 $\pm$ .001 (2) &  \bftab 0.984 $\pm$ .001 (1) \\
Congressional Voting               &         0.971 $\pm$ .019 (4) &      $n>12$ &  0.978 $\pm$ .005 (3) &         1.000 $\pm$ .001 (2) &  \bftab 1.000 $\pm$ .000 (1) \\
Absenteeism                        &         0.778 $\pm$ .050 (4) &      $n>12$ &  0.842 $\pm$ .011 (3) &         0.920 $\pm$ .009 (2) &  \bftab 0.952 $\pm$ .008 (1) \\
Hepatitis                          &         0.931 $\pm$ .038 (4) &      $n>12$ &  0.967 $\pm$ .011 (3) &  \bftab 1.000 $\pm$ .000 (1) &         0.998 $\pm$ .003 (2) \\
German                             &         0.589 $\pm$ .048 (4) &      $n>12$ &  0.891 $\pm$ .010 (3) &  \bftab 0.958 $\pm$ .002 (1) &         0.940 $\pm$ .013 (2) \\
Mushroom                           &  \bftab 1.000 $\pm$ .000 (1) &      $n>12$ &  0.993 $\pm$ .005 (4) &  \bftab 1.000 $\pm$ .000 (1) &  \bftab 1.000 $\pm$ .000 (1) \\
Credit Card                        &         0.651 $\pm$ .020 (4) &      $n>12$ &  0.691 $\pm$ .002 (3) &         0.710 $\pm$ .003 (2) &  \bftab 0.758 $\pm$ .008 (1) \\
Horse Colic                        &         0.882 $\pm$ .033 (4) &      $n>12$ &  0.913 $\pm$ .019 (3) &  \bftab 1.000 $\pm$ .000 (1) &         0.963 $\pm$ .008 (2) \\
Thyroid                            &         0.914 $\pm$ .012 (4) &      $n>12$ &  0.927 $\pm$ .014 (3) &         0.937 $\pm$ .002 (2) &  \bftab 0.987 $\pm$ .002 (1) \\
Cervical Cancer                    &         0.593 $\pm$ .128 (4) &      $n>12$ &  0.691 $\pm$ .046 (3) &         0.767 $\pm$ .020 (2) &  \bftab 0.925 $\pm$ .027 (1) \\
Spambase                           &         0.905 $\pm$ .020 (2) &      $n>12$ &  0.858 $\pm$ .020 (4) &         0.879 $\pm$ .002 (3) &  \bftab 0.965 $\pm$ .003 (1) \\
\midrule
Mean $\uparrow$                            &         0.823 $\pm$ .132 (4) &  -                          &  0.874 $\pm$ .098 (3) &         0.925 $\pm$ .087 (2) &  \bftab 0.938 $\pm$ .065 (1) \\
Mean Reciprocal Rank (MRR) $\uparrow$               &         0.358 $\pm$ .226 (3) &  0.220 $\pm$ .045 (5)      &  0.305 $\pm$ .044 (4) &         0.712 $\pm$ .263 (2) &  \bftab 0.754 $\pm$ .282 (1) \\
\midrule \midrule
Iris                 &  0.978 $\pm$ .017 (2) &  0.926 $\pm$ .030 (5) &  0.947 $\pm$ .008 (4) &  \bftab 0.996 $\pm$ .004 (1) &         0.973 $\pm$ .010 (3) \\
Balance Scale        &  0.793 $\pm$ .034 (2) &  0.602 $\pm$ .142 (5) &  0.665 $\pm$ .034 (4) &         0.700 $\pm$ .022 (3) &  \bftab 0.932 $\pm$ .010 (1) \\
Car                  &  0.807 $\pm$ .048 (4) &  0.681 $\pm$ .087 (5) &  0.863 $\pm$ .055 (3) &         0.885 $\pm$ .029 (2) &  \bftab 0.974 $\pm$ .010 (1) \\
Glass                &  0.785 $\pm$ .045 (4) &  0.804 $\pm$ .020 (3) &  0.699 $\pm$ .049 (5) &         0.837 $\pm$ .009 (2) &  \bftab 0.978 $\pm$ .007 (1) \\
Contraceptive        &  0.545 $\pm$ .067 (5) &  0.591 $\pm$ .105 (4) &  0.723 $\pm$ .030 (3) &         0.822 $\pm$ .016 (2) &  \bftab 0.865 $\pm$ .018 (1) \\
Solar Flare          &  0.833 $\pm$ .090 (3) &  0.729 $\pm$ .080 (5) &  0.794 $\pm$ .123 (4) &         0.904 $\pm$ .047 (2) &  \bftab 0.943 $\pm$ .061 (1) \\
Wine                 &  0.995 $\pm$ .009 (3) &  0.996 $\pm$ .006 (2) &  0.931 $\pm$ .019 (5) &         0.978 $\pm$ .004 (4) &  \bftab 1.000 $\pm$ .000 (1) \\
Zoo                  &  0.990 $\pm$ .029 (3) &      $n>12$ &  0.831 $\pm$ .129 (4) &  \bftab 1.000 $\pm$ .000 (1) &  \bftab 1.000 $\pm$ .000 (2) \\
Lymphography         &  0.974 $\pm$ .018 (2) &      $n>12$ &  0.915 $\pm$ .072 (4) &         0.957 $\pm$ .130 (3) &  \bftab 0.988 $\pm$ .004 (1) \\
Segment              &  0.959 $\pm$ .005 (2) &      $n>12$ &  0.715 $\pm$ .109 (4) &         0.799 $\pm$ .003 (3) &  \bftab 0.987 $\pm$ .002 (1) \\
Dermatology          &  0.965 $\pm$ .014 (2) &      $n>12$ &  0.792 $\pm$ .110 (4) &         0.954 $\pm$ .007 (3) &  \bftab 0.986 $\pm$ .005 (1) \\
Landsat              &  0.824 $\pm$ .010 (2) &      $n>12$ &  0.630 $\pm$ .083 (4) &         0.792 $\pm$ .004 (3) &  \bftab 0.939 $\pm$ .005 (1) \\
Annealing            &  0.885 $\pm$ .089 (4) &      $n>12$ &  0.974 $\pm$ .013 (3) &         0.999 $\pm$ .001 (2) &  \bftab 0.999 $\pm$ .001 (1) \\
Splice               &  0.886 $\pm$ .026 (2) &      $n>12$ &  0.783 $\pm$ .062 (3) &             $>$ 60 min &  \bftab 0.992 $\pm$ .004 (1) \\
\midrule
Mean $\uparrow$              &  0.873 $\pm$ .123 (3) &  -                     &  0.804 $\pm$ .110 (4) &         0.894 $\pm$ .097 (2) &  \bftab 0.968 $\pm$ .038 (1) \\
Mean Reciprocal Rank (MRR) $\uparrow$ &  0.389 $\pm$ .121 (3) &  0.269 $\pm$ .113 (4) &  0.267 $\pm$ .047 (5) &         0.494 $\pm$ .242 (2) &  \bftab 0.952 $\pm$ .178 (1) \\
\bottomrule
\end{tabular}
}
\caption{Train Performance Comparison. We report macro F1-scores (mean $\pm$ stdev over $10$ trials) on the training data. We also report the rank of each approach in brackets. The datasets are sorted by the number of features. The top part comprises binary classification tasks and the bottom part multi-class datasets.}
\label{tab:eval-results-train}
\end{table}

\subsection{Tree Size Comparison}

\begin{table}[H]

\centering
\resizebox{1.0\columnwidth}{!}{
\begin{tabular}{lccccc}
\toprule
{} &                           \multicolumn{2}{l}{Gradient-Based} &            \multicolumn{2}{l}{Non-Greedy} &                       \multicolumn{1}{l}{Greedy} \\ \cmidrule(lr){2-3}  \cmidrule(lr){4-5} \cmidrule(lr){6-6}
  
{} &                           \multicolumn{1}{l}{GradTree (ours)} &                   \multicolumn{1}{l}{DNDT} &            \multicolumn{1}{l}{GeneticTree} &                          \multicolumn{1}{l}{DL8.5 (Optimal)} &                       \multicolumn{1}{l}{CART} \\
\midrule
Blood Transfusion                  &           \phantom{0}29.20 $\pm$ \phantom{00}8.22 (3) &           \phantom{0.0}24.00 $\pm$ \phantom{0.00}0.00 (2) &         \phantom{0}5.20 $\pm$ \phantom{0}3.03 (1) &          24.00 $\pm$ \phantom{0}1.61 (3) &   165.60 $\pm$ \phantom{0}21.84 (5) \\
Banknote Authentication            &          \phantom{0}60.00 $\pm$ \phantom{0}11.22 (5) &           \phantom{0.0}24.00 $\pm$ \phantom{0.00}0.00 (2) &       \bftab 12.80 $\pm$ \phantom{0}5.33 (1) &          26.80 $\pm$ \phantom{0}0.60 (3) &           \phantom{0}44.80 $\pm$ \phantom{00}4.69 (4) \\
Titanic                            &          \phantom{0}39.60 $\pm$ \phantom{0}10.17 (3) &          \phantom{0.}142.00 $\pm$ \phantom{0.00}0.00 (5) & \bftab 10.40 $\pm$ \phantom{0}3.11 (1) &          28.20 $\pm$ \phantom{0}0.98 (3) &           \phantom{0}21.60 $\pm$ \phantom{00}0.92 (2) \\
Raisins                            &         114.80 $\pm$ \phantom{0}33.69 (4) &          \phantom{0.}142.00 $\pm$ \phantom{0.00}0.00 (5) &   \phantom{0}3.60 $\pm$ \phantom{0}1.80 (2) &          29.40 $\pm$ \phantom{0}1.50 (3) &           \bftab \phantom{00}3.00 $\pm$ \phantom{00}0.00 (1) \\
Rice                               &          \phantom{0}41.00 $\pm$ \phantom{0}10.05 (4) &          \phantom{0.}142.00 $\pm$ \phantom{0.00}0.00 (5) &   \phantom{0}3.00 $\pm$ \phantom{0}0.00 (2) &          30.20 $\pm$ \phantom{0}0.98 (3) &           \bftab \phantom{00}3.00 $\pm$ \phantom{00}0.00 (1) \\
Echocardiogram                     &          \phantom{0}43.20 $\pm$ \phantom{0}12.82 (4) &          \phantom{0.}272.00 $\pm$ \phantom{0.00}0.00 (5) &  \bftab \phantom{0}9.00 $\pm$ \phantom{0}3.10 (1) &          30.00 $\pm$ \phantom{0}1.34 (2) &           \phantom{0}36.20 $\pm$ \phantom{00}5.81 (3) \\
Wisconsin Breast Cancer &          \phantom{0}61.80 $\pm$ \phantom{0}13.36 (4) &         1,044.00 $\pm$ \phantom{0.00}0.00 (5) &   \phantom{0}4.00 $\pm$ \phantom{0}1.61 (2) &          29.40 $\pm$ \phantom{0}0.80 (3) &           \bftab \phantom{00}3.00 $\pm$ \phantom{00}0.00 (1) \\
Loan House                         &           \phantom{0}19.20 $\pm$ \phantom{00}5.25 (2) &         2,070.00 $\pm$ \phantom{0.00}0.00 (5) &  \bftab \phantom{0}6.80 $\pm$ \phantom{0}2.75 (1) &          28.20 $\pm$ \phantom{0}1.33 (4) &           \phantom{0}26.20 $\pm$ \phantom{00}2.04 (3) \\
Heart Failure                      &           \phantom{0}27.20 $\pm$ \phantom{00}9.44 (3) &         4,120.00 $\pm$ \phantom{0.00}0.00 (5) &  \bftab \phantom{0}3.20 $\pm$ \phantom{0}0.60 (1) &          30.80 $\pm$ \phantom{0}0.60 (4) &           \phantom{0}14.20 $\pm$ \phantom{00}0.98 (2) \\
Heart Disease                      &           \phantom{0}36.80 $\pm$ \phantom{00}7.45 (4) &                           $n>12$ & \bftab 11.00 $\pm$ \phantom{0}4.10 (1) &          27.60 $\pm$ \phantom{0}1.56 (3) &           \phantom{0}24.60 $\pm$ \phantom{00}3.07 (2) \\
Adult                              &         128.00 $\pm$ \phantom{0}37.19 (3) &                           $n>12$ &  \bftab \phantom{0}9.40 $\pm$ \phantom{0}4.96 (1) &          25.20 $\pm$ \phantom{0}0.60 (2) &         156.60 $\pm$ \phantom{0}23.08 (4) \\
Bank Marketing                     &            \phantom{00}6.80 $\pm$ \phantom{00}3.03 (2) &                           $n>12$ &  \bftab \phantom{0}3.40 $\pm$ \phantom{0}1.20 (1) &          27.20 $\pm$ \phantom{0}2.27 (3) &         110.60 $\pm$ \phantom{0}10.50 (4) \\
Congressional Voting               &            \phantom{00}5.60 $\pm$ \phantom{00}2.38 (2) &                           $n>12$ &  \bftab \phantom{0}5.60 $\pm$ \phantom{0}0.92 (1) &         20.40 $\pm$ \phantom{0}6.70 (4) &           \phantom{0}19.20 $\pm$ \phantom{00}6.42 (3) \\
Absenteeism                        &         148.80 $\pm$ \phantom{0}18.19 (4) &                           $n>12$ &  \bftab \phantom{0}7.80 $\pm$ \phantom{0}0.98 (1) &          30.60 $\pm$ \phantom{0}0.80 (2) &           \phantom{0}43.20 $\pm$ \phantom{00}2.75 (3) \\
Hepatitis                          &           \phantom{0}12.20 $\pm$ \phantom{00}3.60 (2) &                           $n>12$ &  \bftab \phantom{0}5.80 $\pm$ \phantom{0}2.71 (1) &          14.80 $\pm$ \phantom{0}2.09 (4) &           \phantom{0}12.60 $\pm$ \phantom{00}1.74 (3) \\
German                             &           \phantom{0}27.60 $\pm$ \phantom{00}4.10 (2) &                           $n>12$ &  \bftab \phantom{0}6.20 $\pm$ \phantom{0}0.98 (1) &          30.40 $\pm$ \phantom{0}0.92 (3) &           \phantom{0}34.60 $\pm$ \phantom{00}5.71 (4) \\
Mushroom                           &           \phantom{0}25.60 $\pm$ \phantom{00}5.73 (4) &                           $n>12$ &  \bftab \phantom{0}5.20 $\pm$ \phantom{0}1.66 (1) &          21.20 $\pm$ \phantom{0}2.75 (3) &           \phantom{0}14.00 $\pm$ \phantom{00}1.34 (2) \\
Credit Card                        &          \phantom{0}92.60 $\pm$ \phantom{0}39.64 (3) &                           $n>12$ &  \bftab \phantom{0}3.00 $\pm$ \phantom{0}0.00 (1) &          31.00 $\pm$ \phantom{0}0.00 (2) &         354.80 $\pm$ \phantom{0}31.60 (4) \\
Horse Colic                        &           \phantom{0}22.40 $\pm$ \phantom{00}4.74 (3) &                           $n>12$ & \bftab  \phantom{0}4.00 $\pm$ \phantom{0}1.61 (1) &          29.40 $\pm$ \phantom{0}1.20 (4) &           \phantom{0}12.60 $\pm$ \phantom{00}1.20 (2) \\
Thyroid                            &          \phantom{0}96.80 $\pm$ \phantom{0}16.86 (4) &                           $n>12$ &  \bftab \phantom{0}5.60 $\pm$ \phantom{0}1.80 (1) &          30.60 $\pm$ \phantom{0}0.80 (2) &           \phantom{0}66.40 $\pm$ \phantom{00}4.65 (3) \\
Cervical Cancer                    &          \phantom{0}34.00 $\pm$ \phantom{0}23.67 (3) &                           $n>12$ & \bftab 14.00 $\pm$ \phantom{0}7.55 (1) &          30.80 $\pm$ \phantom{0}0.60 (2) &           \phantom{0}99.00 $\pm$ \phantom{00}8.53 (4) \\
Spambase                           &         120.20 $\pm$ \phantom{0}37.01 (3) &                           $n>12$ & \bftab 12.40 $\pm$ \phantom{0}5.80 (1) &          29.600 $\pm$ \phantom{0}0.917 (2) &         201.00 $\pm$ \phantom{0}14.64 (4) \\
\midrule
Mean $\uparrow$                            &            \phantom{0}54.245 $\pm$ \phantom{0}42.82 (3) &    \phantom{0.}886.67 $\pm$ 1,387.86 (5) &       \bftab  \phantom{0}6.88 $\pm$ \phantom{0}3.46 (1) &           27.54 $\pm$ \phantom{0}4.15 (2) &            \phantom{0}66.67 $\pm$ \phantom{0}86.22 (4) \\
Mean Reciprocal Rank (MRR) $\uparrow$               &            \phantom{00}0.33 $\pm$ \phantom{00}0.10 (4) &      \phantom{0.00}0.27 $\pm$ \phantom{0.00}0.13 (5) &       \bftab  \phantom{0}0.93 $\pm$ \phantom{0}0.18 (1) &           \phantom{0}0.37 $\pm$ \phantom{0}0.10 (3) &            \phantom{00}0.43 $\pm$ \phantom{00}0.25 (2) \\
\midrule \midrule
Iris                 &           \phantom{0}14.40 $\pm$ \phantom{00}2.54 (3) &           \phantom{0.0}24.00 $\pm$ \phantom{0.00}0.00 (4) &  \bftab  9.00 $\pm$ \phantom{0}3.10 (1) &         29.20 $\pm$ \phantom{0}1.40 (5) &            \phantom{00}9.40 $\pm$ \phantom{00}1.50 (2) \\
Balance Scale        &         118.60 $\pm$ \phantom{0}27.10 (4) &         \bftab  \phantom{0.0}24.00 $\pm$ \phantom{0.00}0.00 (1) &  25.40 $\pm$ 11.31 (2) &         31.00 $\pm$ \phantom{0}0.00 (3) &          161.20 $\pm$ \phantom{00}8.92 (5) \\
Car                  &          \phantom{0}47.00 $\pm$ \phantom{0}13.57 (3) &           \phantom{0.0}76.00 $\pm$ \phantom{0.00}0.00 (5) & \bftab 32.60 $\pm$ 21.35 (1) &         29.80 $\pm$ \phantom{0}0.98 (2) &          \phantom{0}57.00 $\pm$ \phantom{0}13.02 (4) \\
Glass                &          \phantom{0}56.80 $\pm$ \phantom{0}11.01 (3) &          \phantom{0.}530.00 $\pm$ \phantom{0.00}0.00 (5) & \bftab 21.80 $\pm$ 11.53 (1) &         30.60 $\pm$ \phantom{0}0.80 (2) &           \phantom{0}68.00 $\pm$ \phantom{00}6.34 (4) \\
Contraceptive        &          \phantom{0}40.60 $\pm$ \phantom{0}11.48 (3) &          \phantom{0.}530.00 $\pm$ \phantom{0.00}0.00 (5) &  \bftab 12.80 $\pm$ \phantom{0}3.84 (1) &         30.00 $\pm$ \phantom{0}1.34 (2) &           \phantom{0}48.60 $\pm$ \phantom{00}7.14 (4) \\
Solar Flare          &           \phantom{0}49.80 $\pm$ \phantom{00}9.30 (4) &         1,044.00 $\pm$ \phantom{0.00}0.00 (5) &  38.40 $\pm$ 21.41 (3) &         26.80 $\pm$ \phantom{0}3.28 (1) &           \phantom{0}32.00 $\pm$ \phantom{00}3.13 (2) \\
Wine                 &           \phantom{0}24.00 $\pm$ \phantom{00}4.12 (3) &         4,120.00 $\pm$ \phantom{0.00}0.00 (5) &  \bftab 11.20 $\pm$ \phantom{0}4.77 (1) &         28.20 $\pm$ \phantom{0}1.60 (4) &          \phantom{0}13.40 $\pm$ \phantom{00}2.33 (2) \\
Zoo                  &           \phantom{0}25.00 $\pm$ \phantom{00}3.90 (3) &                           $n>12$ &  32.60 $\pm$ 19.30 (4) &         24.60 $\pm$ \phantom{0}0.80 (2) &          \bftab \phantom{0}18.00 $\pm$ \phantom{00}1.34 (1) \\
Lymphography         &           \phantom{0}36.20 $\pm$ \phantom{00}6.34 (4) &                           $n>12$ &  \bftab 14.40 $\pm$ \phantom{0}6.82 (1) &         27.00 $\pm$ \phantom{0}1.27 (3) &           \phantom{0}16.80 $\pm$ \phantom{00}1.89 (2) \\
Segment              &         133.00 $\pm$ \phantom{0}21.85 (4) &                           $n>12$ & \bftab 24.40 $\pm$ 11.42 (1) &         31.00 $\pm$ \phantom{0}0.00 (2) &           \phantom{0}79.60 $\pm$ \phantom{00}3.80 (3) \\
Dermatology          &           \phantom{0}33.60 $\pm$ \phantom{00}7.27 (4) &                           $n>12$ &  \bftab 20.00 $\pm$ \phantom{0}7.76 (1) &         28.00 $\pm$ \phantom{0}1.84 (2) &           \phantom{0}28.00 $\pm$ \phantom{00}1.84 (2) \\
Landsat              &         360.60 $\pm$ \phantom{0}21.12 (3) &                           $n>12$ &  \bftab 16.00 $\pm$ \phantom{0}7.34 (1) &         31.00 $\pm$ \phantom{0}0.00 (2) &         425.60 $\pm$ \phantom{0}18.53 (4) \\
Annealing            &           \phantom{0}19.60 $\pm$ \phantom{00}2.84 (2) &                           $n>12$ &  \bftab 12.60 $\pm$ \phantom{0}3.32 (1) &         28.40 $\pm$ \phantom{0}2.01 (4) &           \phantom{0}24.40 $\pm$ \phantom{00}2.01 (3) \\
Splice               &         247.40 $\pm$ \phantom{0}53.00 (3) &                           $n>12$ &  \bftab 11.80 $\pm$ \phantom{0}6.40 (1) &                         $n>12$ &          \phantom{0}81.40 $\pm$ \phantom{0}10.69 (2) \\
\midrule
Mean $\uparrow$              &            \phantom{0}86.19 $\pm$ 101.07 (4) &      \phantom{0.}906.86 $\pm$ 1,464.94 (5) &  \bftab 20.21 $\pm$ 9.31 (1) &          28.95 $\pm$ \phantom{0}1.98 (2) &            \phantom{0}75.96 $\pm$ 108.42 (3) \\
Mean Reciprocal Rank (MRR) $\uparrow$ &            \phantom{00}0.32 $\pm$ \phantom{00}0.07 (5) &      \phantom{0.00}0.32 $\pm$ \phantom{0.00}0.30 (4) &  \bftab  \phantom{0}0.86 $\pm$ 0.28 (1) &          \phantom{0}0.45 $\pm$ \phantom{0}0.20 (2) &            \phantom{00}0.42 $\pm$ \phantom{00}0.21 (3) \\
\bottomrule
\end{tabular}
}
\caption{Tree Size Comparison. We report the average tree size based on the optimized hyperparameters (mean $\pm$ stdev over $10$ trials). We also report the rank of each approach in brackets. The datasets are sorted by the number of features. The top part comprises binary classification tasks and the bottom part multi-class datasets.}
\label{tab:eval-results-tree-size}
\end{table}

\subsection{Runtime Comparison}

\begin{table}[H]

\centering
\resizebox{1.0\columnwidth}{!}{
\begin{tabular}{lccccc}
\toprule
{} &                           \multicolumn{2}{l}{Gradient-Based} &            \multicolumn{2}{l}{Non-Greedy} &                       \multicolumn{1}{l}{Greedy} \\ \cmidrule(lr){2-3}  \cmidrule(lr){4-5} \cmidrule(lr){6-6}
  
{} &                           \multicolumn{1}{l}{GradTree (ours)} &                   \multicolumn{1}{l}{DNDT} &            \multicolumn{1}{l}{GeneticTree} &                          \multicolumn{1}{l}{DL8.5 (Optimal)} &                       \multicolumn{1}{l}{CART} \\
\midrule
Blood Transfusion                  &        \phantom{0}13.01 $\pm$ \phantom{0}1.00 (5) &    \phantom{0}3.70 $\pm$ \phantom{0}1.00 (3) &   \phantom{0}10.51 $\pm$ \phantom{0}4.00 (4) &    \phantom{00}0.03 $\pm$ 0.00 (2) &  \bftab 0.00 $\pm$ 0.00 (1) \\
Banknote Authentication            &        \phantom{0}13.21 $\pm$ \phantom{0}2.00 (4) &   34.43 $\pm$ \phantom{0}1.00 (5) &   \phantom{0}10.32 $\pm$ \phantom{0}3.00 (3) &    \phantom{00}0.03 $\pm$ 0.00 (2) &  \bftab 0.00 $\pm$ 0.00 (1) \\
Titanic                            &        \phantom{0}29.58 $\pm$ \phantom{0}1.00 (5) &    \phantom{0}6.82 $\pm$ \phantom{0}0.00 (4) &    \phantom{00}3.68 $\pm$ \phantom{0}1.00 (3) &    \phantom{00}0.26 $\pm$ 0.00 (2) &  \bftab 0.00 $\pm$ 0.00 (1) \\
Raisins                            &        \phantom{0}30.61 $\pm$ \phantom{0}2.00 (5) &    \phantom{0}8.50 $\pm$ \phantom{0}1.00 (4) &    \phantom{00}0.62 $\pm$ \phantom{0}0.00 (3) &    \phantom{00}0.20 $\pm$ 0.00 (2) &  \bftab 0.00 $\pm$ 0.00 (1) \\
Rice                               &        \phantom{0}21.49 $\pm$ \phantom{0}4.00 (5) &   11.96 $\pm$ \phantom{0}1.00 (4) &    \phantom{00}1.33 $\pm$ \phantom{0}0.00 (3) &    \phantom{00}0.48 $\pm$ 0.00 (2) &  \bftab 0.01 $\pm$ 0.00 (1) \\
Echocardiogram                     &         \phantom{00}9.04 $\pm$ \phantom{0}1.00 (4) &    \phantom{0}9.56 $\pm$ \phantom{0}2.00 (5) &    \phantom{00}0.30 $\pm$ \phantom{0}0.00 (3) &    \phantom{00}0.09 $\pm$ 0.00 (2) &  \bftab 0.00 $\pm$ 0.00 (1) \\
Wisconsin Breast Cancer &        \phantom{0}29.25 $\pm$ \phantom{0}2.00 (5) &    \phantom{0}6.84 $\pm$ \phantom{0}1.00 (4) &    \phantom{00}1.19 $\pm$ \phantom{0}0.00 (3) &    \phantom{00}0.46 $\pm$ 0.00 (2) &  \bftab 0.00 $\pm$ 0.00 (1) \\
Loan House                         &        \phantom{0}10.80 $\pm$ \phantom{0}1.00 (4) &   52.79 $\pm$ \phantom{0}7.00 (5) &    \phantom{00}6.75 $\pm$ \phantom{0}2.00 (3) &    \phantom{00}0.52 $\pm$ 0.00 (2) &  \bftab 0.00 $\pm$ 0.00 (1) \\
Heart Failure                      &        \phantom{0}28.40 $\pm$ \phantom{0}1.00 (5) &  23.75 $\pm$ 10.00 (4) &    \phantom{00}1.47 $\pm$ \phantom{0}0.00 (3) &    \phantom{00}0.20 $\pm$ 0.00 (2) &  \bftab 0.00 $\pm$ 0.00 (1) \\
Heart Disease                      &        \phantom{0}15.57 $\pm$ \phantom{0}2.00 (4) &        $n>12$ &    \phantom{00}4.45 $\pm$ \phantom{0}1.00 (3) &    \phantom{00}0.64 $\pm$ 0.00 (2) &  \bftab 0.00 $\pm$ 0.00 (1) \\
Adult                              &       \phantom{0}86.24 $\pm$ 12.00 (4) &        $n>12$ &   \phantom{0}26.01 $\pm$ \phantom{0}7.00 (3) &   \phantom{0}22.32 $\pm$ 1.00 (2) &  \bftab 0.07 $\pm$ 0.00 (1) \\
Bank Marketing                     &      153.49 $\pm$ 38.00 (4) &        $n>12$ &  129.43 $\pm$ \phantom{0}8.00 (3) &   \phantom{0}54.55 $\pm$ 1.00 (2) &  \bftab 0.07 $\pm$ 0.00 (1) \\
Congressional Voting               &        \phantom{0}32.94 $\pm$ \phantom{0}3.00 (4) &        $n>12$ &    \phantom{00}6.50 $\pm$ \phantom{0}1.00 (3) &    \phantom{00}4.10 $\pm$ 7.00 (2) &  \bftab 0.00 $\pm$ 0.00 (1) \\
Absenteeism                        &        \phantom{0}28.94 $\pm$ \phantom{0}1.00 (4) &        $n>12$ &   \phantom{0}14.05 $\pm$ \phantom{0}3.00 (3) &    \phantom{00}3.89 $\pm$ 0.00 (2) &  \bftab 0.00 $\pm$ 0.00 (1) \\
Hepatitis                          &        \phantom{0}28.43 $\pm$ \phantom{0}1.00 (4) &        $n>12$ &    \phantom{00}2.27 $\pm$ \phantom{0}1.00 (3) &    \phantom{00}0.02 $\pm$ 0.00 (2) &  \bftab 0.00 $\pm$ 0.00 (1) \\
German                             &        \phantom{0}12.31 $\pm$ \phantom{0}2.00 (4) &        $n>12$ &    \phantom{00}3.14 $\pm$ \phantom{0}1.00 (2) &   \phantom{0}10.29 $\pm$ 0.00 (3) &  \bftab 0.00 $\pm$ 0.00 (1) \\
Mushroom                           &       \phantom{0}58.81 $\pm$ 70.00 (4) &        $n>12$ &   \phantom{0}12.51 $\pm$ \phantom{0}2.00 (3) &    \phantom{00}0.51 $\pm$ 0.00 (2) &  \bftab 0.01 $\pm$ 0.00 (1) \\
Credit Card                        &      111.43 $\pm$ 36.00 (3) &        $n>12$ &    \phantom{00}4.42 $\pm$ \phantom{0}0.00 (2) &  298.57 $\pm$ 6.00 (4) &  \bftab 0.35 $\pm$ 0.00 (1) \\
Horse Colic                        &         \phantom{00}9.81 $\pm$ \phantom{0}1.00 (4) &        $n>12$ &    \phantom{00}0.32 $\pm$ \phantom{0}0.00 (2) &    \phantom{00}2.37 $\pm$ 2.00 (3) &  \bftab 0.00 $\pm$ 0.00 (1) \\
Thyroid                            &       \phantom{0}35.97 $\pm$ 13.00 (2) &        $n>12$ &  \phantom{0}38.29 $\pm$ 15.00 (3) &  109.12 $\pm$ 3.00 (4) &  \bftab 0.01 $\pm$ 0.00 (1) \\
Cervical Cancer                    &        \phantom{0}13.47 $\pm$ \phantom{0}2.00 (4) &        $n>12$ &    \phantom{00}2.76 $\pm$ \phantom{0}1.00 (3) &    \phantom{00}0.13 $\pm$ 0.00 (2) &  \bftab 0.01 $\pm$ 0.00 (1) \\
Spambase                           &        \phantom{0}41.98 $\pm$ \phantom{0}5.00 (4) &        $n>12$ &   \phantom{0}19.29 $\pm$ \phantom{0}6.00 (3) &    \phantom{00}4.58 $\pm$ 0.00 (2) &  \bftab 0.04 $\pm$ 0.00 (1) \\
\midrule
Mean $\uparrow$                          &   44.75 $\pm$ 38.66 (4) &    -                  &    13.62 $\pm$ 27.54 (2) &   \phantom{0}23.33 $\pm$ 66.45 (3) &       \bftab  0.03 $\pm$ 0.08 (1) \\
Mean Reciprocal Rank (MRR) $\uparrow$             &  \phantom{0}0.24 $\pm$ \phantom{0}0.04 (4) &    \phantom{0}0.24 $\pm$ \phantom{0}0.042 (5) &    \phantom{0}0.35 $\pm$ \phantom{0}0.06 (3) &    \phantom{00}0.46 $\pm$ \phantom{0}0.08 (2) &        \bftab 1.00 $\pm$ 0.00 (1) \\
\midrule \midrule
Iris                 &         \phantom{0}7.25 $\pm$ \phantom{0}1.00 (5) &    \phantom{0}2.96 $\pm$ \phantom{0}0.00 (4) &    \phantom{0}0.97 $\pm$ \phantom{0}0.00 (3) &       \phantom{00}0.02 $\pm$ \phantom{0}0.00 (2) &  \bftab 0.00 $\pm$ 0.00 (1) \\
Balance Scale        &        11.61 $\pm$ \phantom{0}2.00 (5) &   10.12 $\pm$ \phantom{0}9.00 (4) &    \phantom{0}8.18 $\pm$ \phantom{0}1.00 (3) &       \phantom{00}0.04 $\pm$ \phantom{0}0.00 (2) &  \bftab 0.00 $\pm$ 0.00 (1) \\
Car                  &        23.72 $\pm$ \phantom{0}3.00 (3) &  46.91 $\pm$ 43.00 (5) &  29.69 $\pm$ 10.00 (4) &       \phantom{00}0.24 $\pm$ \phantom{0}0.00 (2) &  \bftab 0.01 $\pm$ 0.00 (1) \\
Glass                &        30.14 $\pm$ \phantom{0}2.00 (5) &    \phantom{0}5.49 $\pm$ \phantom{0}0.00 (3) &   12.98 $\pm$ \phantom{0}5.00 (4) &       \phantom{00}0.18 $\pm$ \phantom{0}0.00 (2) &  \bftab 0.00 $\pm$ 0.00 (1) \\
Contraceptive        &        13.17 $\pm$ \phantom{0}2.00 (5) &   12.67 $\pm$ \phantom{0}2.00 (4) &    \phantom{0}3.95 $\pm$ \phantom{0}1.00 (3) &       \phantom{00}0.26 $\pm$ \phantom{0}0.00 (2) &  \bftab 0.00 $\pm$ 0.00 (1) \\
Solar Flare          &        22.11 $\pm$ \phantom{0}2.00 (4) &   25.65 $\pm$ \phantom{0}3.00 (5) &   14.73 $\pm$ \phantom{0}2.00 (3) &       \phantom{00}0.54 $\pm$ \phantom{0}0.00 (2) &  \bftab 0.00 $\pm$ 0.00 (1) \\
Wine                 &        30.10 $\pm$ \phantom{0}1.00 (4) &  58.49 $\pm$ 10.00 (5) &    \phantom{0}0.13 $\pm$ \phantom{0}0.00 (2) &       \phantom{00}0.54 $\pm$ \phantom{0}0.00 (3) &  \bftab 0.00 $\pm$ 0.00 (1) \\
Zoo                  &        19.26 $\pm$ \phantom{0}7.00 (4) &        $n>12$ &    \phantom{0}9.56 $\pm$ \phantom{0}3.00 (3) &       \phantom{00}0.02 $\pm$ \phantom{0}0.00 (2) &  \bftab 0.00 $\pm$ 0.00 (1) \\
Lymphography         &         \phantom{0}7.89 $\pm$ \phantom{0}1.00 (4) &        $n>12$ &    \phantom{0}4.36 $\pm$ \phantom{0}1.00 (3) &       \phantom{00}0.14 $\pm$ \phantom{0}0.00 (2) &  \bftab 0.00 $\pm$ 0.00 (1) \\
Segment              &        16.18 $\pm$ \phantom{0}3.00 (4) &        $n>12$ &    \phantom{0}2.97 $\pm$ \phantom{0}1.00 (2) &      \phantom{0}14.32 $\pm$ \phantom{0}1.00 (3) &  \bftab 0.02 $\pm$ 0.00 (1) \\
Dermatology          &         \phantom{0}9.91 $\pm$ \phantom{0}3.00 (3) &        $n>12$ &    \phantom{0}1.72 $\pm$ \phantom{0}1.00 (2) &      \phantom{0}20.71 $\pm$ \phantom{0}6.00 (4) &  \bftab 0.00 $\pm$ 0.00 (1) \\
Landsat              &       38.13 $\pm$ 11.00 (3) &        $n>12$ &  33.83 $\pm$ 11.00 (2) &     830.06 $\pm$ \phantom{0}9.00 (4) &  \bftab 0.05 $\pm$ 0.00 (1) \\
Annealing            &        36.19 $\pm$ \phantom{0}3.00 (3) &        $n>12$ &    \phantom{0}1.36 $\pm$ \phantom{0}0.00 (2) &    104.57 $\pm$ 24.00 (4) &  \bftab 0.01 $\pm$ 0.00 (1) \\
Splice               &        19.20 $\pm$ \phantom{0}2.00 (3) &        $n>12$ &    \phantom{0}2.88 $\pm$ \phantom{0}1.00 (2) &  $>$ 60 min &  \bftab 0.02 $\pm$ 0.00 (1) \\
\midrule
Mean $\uparrow$              &   20.35 $\pm$ 10.21 (3) &    - &    \phantom{0}9.09 $\pm$ 10.63 (2) &   \phantom{0}74.74 $\pm$ 228.75     (4) &         \bftab  0.01 $\pm$ 0.01 (1) \\
Mean Reciprocal Rank (MRR) $\uparrow$ &   \phantom{0}0.27 $\pm$ \phantom{0}0.06 (4) &    \phantom{0}0.24 $\pm$ \phantom{0}0.05 (5) &    \phantom{0}0.39 $\pm$ \phantom{0}0.10 (2) &       \phantom{00}0.38 $\pm$ \phantom{00}0.13 (3) &         \bftab 1.00 $\pm$ 0.00 (1) \\
\bottomrule
\end{tabular}
}
\caption{Runtime Comparison. We report runtime without restarts based on the optimized hyperparameters (mean $\pm$ stdev over $10$ trials). We also report the rank of each approach in brackets. The datasets are sorted by the number of features. The top part comprises binary classification tasks and the bottom part multi-class datasets.}
\label{tab:eval-results-runtime}
\end{table}
\newpage

\section{Hyperparameters}\label{A:hyperparams}
In the following, we report the hyperparameters used for each approach. The hyperparameters were selected based on a random search over a predefined parameter range for GradTree, CART and GeneticTree and are summarized in Table~\ref{tab:hpos_GradTree} to Table~\ref{tab:hpos_sklearn}. All parameters that were considered are noted in the tables. The number of trials ($300$) was equal for each approach. For GradTree and DNDT, we did not optimize the batch size as well as the number of epochs, but used early stopping with a predefined patience of $200$. Additionally, we used $3$ random restarts to prevent bad initial parametrizations and selected the best model based on the validation loss.
A gradient-based optimization allows using an arbitrary loss function for the optimization. During preliminary experiments, we observed that it is beneficial to adjust the loss function for specific datasets. More specifically, we allowed adjusting the cross-entropy loss by adding a focal factor~\citep{lin2017focal} of $3$. Additionally, we considered using PolyLoss~\citep{leng2022polyloss} within the HPO to tailor the loss function for the specific task. 
For GeneticTree, we used a complexity penalty as it is commonly used in genetic algorithms. This has a strong impact on the tree size of the learned DT and can explain the small complexity of GeneticTree. To ensure that this complexity penalty does have a significant impact on the performance of the method, we conducted an additional experiment comparing GeneticTree with and without complexity penalty (see Table~\ref{tab:genetictree_comparison}). Removing the complexity penalty in the genetic algorithm resulted in a small increase of the average performance to $0.713$ (GradTree $0.758$) for binary and $0.573$ (GradTree $0.619$) for multi-class. However, removing the complexity penalty also led to a significant increase in the tree size (e.g. from 7 to 1,600 for binary classification). Thus, interpretability of the models is substantially reduced and runtimes substantially increased. Most importantly, the change in the performance does not impact the overall results: When excluding the complexity penalty of the genetic benchmark completely, the relative performance of the methods (number of wins and rank) remained unchanged.
For DL8.5 the relevant tunable hyperparameters according to the authors are the maximum depth and the minimum support. However, the maximum depth strongly impacts the runtime, which is why we fixed the maximum depth to $4$, similar to the maximum depth used during the experiments of \citet{murtree} and \citet{aglin2020learning}. Running the experiments with a higher depth becomes infeasible for many datasets. In preliminary experiments, we also observed that changing the depth does not have a positive impact on the performance (e.g. increasing the depth to $5$ results in a decrease in the test performance). Similarly, reducing the depth resulted in a reduced performance. Furthermore, to assure a fair comparison, we fixed the minimum support to $1$ which is equal to the pruning of GradTree. Additionally, we observed in our preliminary experiments, that increasing the minimum support reduces overfitting, but has no positive impact on the test performance (i.e. the train performance decreases, but the test performance is not improved).
For DNDT, the number of cut points is the tunable hyperparameter of the model, according to \citet{dndt}. However, it has to be restricted to $1$ in order to generate binary trees for comparability reasons. Therefore, we only optimized the temperature and the learning rate.
Furthermore, we extended their implementation to use early stopping based on the validation loss, similar to GradTree, to reduce the runtime and prevent overfitting. Further details can be found in our implementation.

\begin{table}[b]
    \centering

    \begin{tabular}{lrrrr}
         \toprule
         & \multicolumn{2}{l}{Binary} & \multicolumn{2}{l}{Multi} \\ \cmidrule(lr){2-3} \cmidrule(lr){4-5}
         & \multicolumn{1}{l}{Penalty} & \multicolumn{1}{l}{No Penalty} & \multicolumn{1}{l}{Penalty} & \multicolumn{1}{l}{No Penalty} \\\midrule
         F1-Score   &  0.671 & 0.713 &  0.546 & 0.573 \\
         Tree Size  &  \phantom{0,00}7 & 1,600 &  \phantom{0,0}20 & 2,201  \\
         \bottomrule
    \end{tabular}
\caption{GeneticTree Comparison. We compare GeneticTree with and without complexity penalty.}
\label{tab:genetictree_comparison}
    
\end{table}


\begin{table}[H]
\centering
\resizebox{\columnwidth}{!}{
\begin{tabular}{lrrrrlllr}
\toprule
 Dataset Name &  depth &  lr\_index &  lr\_values &  lr\_leaf & index\_activation &                      loss &  polyLoss &  polyLossEpsilon \\
\midrule
                 Blood Transfusion &      8 &      0.010 &       0.100 &     0.010 &                   entmax &              crossentropy &     False &                2 \\
           Banknote Authentication &      7 &      0.050 &       0.050 &     0.100 &                   entmax & focal\_crossentropy &      True &                2 \\
                           Titanic &     10 &      0.005 &       0.010 &     0.010 &                   entmax &              crossentropy &     False &                2 \\
                           Raisins &     10 &      0.005 &       0.005 &     0.100 &                   entmax &              crossentropy &      True &                5 \\
                              Rice &      7 &      0.050 &       0.010 &     0.010 &                   entmax &              crossentropy &     False &                2 \\
                    Echocardiogram &      8 &      0.010 &       0.050 &     0.100 &                   entmax &              crossentropy &      True &                5 \\
Wisconsin Diagnostic Breast Cancer &     10 &      0.050 &       0.010 &     0.100 &                   entmax &              crossentropy &      True &                2 \\
                        Loan House &      8 &      0.005 &       0.100 &     0.010 &                   entmax & focal\_crossentropy &     False &                2 \\
                     Heart Failure &     10 &      0.005 &       0.250 &     0.100 &                   entmax &              crossentropy &     False &                2 \\
                     Heart Disease &      9 &      0.010 &       0.050 &     0.005 &                   entmax & focal\_crossentropy &     False &                2 \\
                             Adult &      8 &      0.050 &       0.005 &     0.050 &                   entmax &              crossentropy &      True &                5 \\
                    Bank Marketing &      8 &      0.250 &       0.250 &     0.050 &                   entmax &              crossentropy &     False &                2 \\
                   Cervical Cancer &      8 &      0.005 &       0.010 &     0.100 &                   entmax &              crossentropy &      True &                2 \\
              Congressional Voting &     10 &      0.005 &       0.050 &     0.010 &                   entmax & focal\_crossentropy &      True &                5 \\
                       Absenteeism &     10 &      0.050 &       0.010 &     0.050 &                   entmax & focal\_crossentropy &      True &                5 \\
                         Hepatitis &     10 &      0.005 &       0.050 &     0.010 &                   entmax & focal\_crossentropy &      True &                5 \\
                            German &      7 &      0.005 &       0.050 &     0.010 &                   entmax &              crossentropy &      True &                2 \\
                          Mushroom &      9 &      0.010 &       0.010 &     0.050 &                   entmax & focal\_crossentropy &     False &                2 \\
                       Credit Card &      8 &      0.050 &       0.100 &     0.010 &                   entmax & focal\_crossentropy &     False &                2 \\
                       Horse Colic &      8 &      0.250 &       0.250 &     0.010 &                   entmax & focal\_crossentropy &     False &                2 \\
                           Thyroid &      8 &      0.010 &       0.010 &     0.050 &                   entmax &              crossentropy &     False &                2 \\
                          Spambase &     10 &      0.005 &       0.010 &     0.010 &                   entmax &              crossentropy &     False &                2 \\
\midrule
         Iris &      7 &      0.005 &       0.005 &      0.05 &                   entmax &              crossentropy &     False &                2 \\
Balance Scale &      8 &      0.050 &       0.010 &      0.10 &                   entmax &              crossentropy &      True &                5 \\
          Car &      9 &      0.010 &       0.010 &      0.01 &                   entmax & focal\_crossentropy &     False &                2 \\
        Glass &     10 &      0.050 &       0.050 &      0.05 &                   entmax & focal\_crossentropy &      True &                5 \\
Contraceptive &      7 &      0.010 &       0.050 &      0.01 &                   entmax &              crossentropy &     False &                2 \\
  Solar Flare &      8 &      0.005 &       0.010 &      0.20 &                   entmax & focal\_crossentropy &      True &                2 \\
         Wine &     10 &      0.010 &       0.050 &      0.01 &                   entmax & focal\_crossentropy &     False &                2 \\
          Zoo &      9 &      0.050 &       0.010 &      0.10 &                   entmax & focal\_crossentropy &      True &                2 \\
 Lymphography &      8 &      0.050 &       0.010 &      0.05 &                   entmax &              crossentropy &      True &                2 \\
      Segment &      7 &      0.005 &       0.005 &      0.05 &                   entmax &              crossentropy &     False &                2 \\
  Dermatology &      7 &      0.010 &       0.010 &      0.10 &                   entmax &              crossentropy &      True &                2 \\
      Landsat &      8 &      0.005 &       0.010 &      0.05 &                   entmax &              crossentropy &      True &                5 \\
    Annealing &     10 &      0.250 &       0.050 &      0.01 &                   entmax &              crossentropy &     False &                2 \\
       Splice &      9 &      0.010 &       0.005 &      0.05 &                   entmax &              crossentropy &     False &                2 \\
\bottomrule
\end{tabular}
}
\caption{GradTree Hyperparameters}

\label{tab:hpos_GradTree}
\end{table}

\begin{table}[H]
\centering
\small
\resizebox{0.6\columnwidth}{!}{
\begin{tabular}{lrrr}
\toprule
                      Dataset Name &  learning\_rate &  temperature & num\_cut  \\
\midrule
                 Blood Transfusion &           0.050 &        0.100 &     1  \\
           Banknote Authentication &           0.001 &        0.100 &     1  \\
                           Titanic &           0.050 &        0.001 &     1  \\
                           Raisins &           0.050 &        0.001 &     1  \\
                              Rice &           0.100 &        0.010 &     1  \\
                    Echocardiogram &           0.005 &        1.000 &     1  \\
Wisconsin Diagnostic Breast Cancer &           0.100 &        0.010 &     1  \\
                        Loan House &           0.001 &        1.000 &     1  \\
                     Heart Failure &           0.005 &        1.000 &     1  \\

\midrule
         Iris &           0.050 &         0.100 &     1  \\
Balance Scale &           0.005 &         0.100 &     1  \\
          Car &           0.010 &         0.010 &     1  \\
        Glass &           0.100 &         0.010 &     1  \\
Contraceptive &           0.001 &         0.010 &     1  \\
  Solar Flare &           0.050 &         0.100 &     1  \\
         Wine &           0.001 &         0.100 &     1  \\
\bottomrule
\end{tabular}
}
\caption{DNDT Hyperparameters}

\label{tab:hpos_dndt}
\end{table}

\begin{table}[H]
\centering
\small
\resizebox{0.8\columnwidth}{!}{
\begin{tabular}{lrrrrr}

\toprule
                      Dataset Name &  n\_thresholds &  n\_trees &  max\_iter &  cross\_prob &  mutation\_prob \\
\midrule
                 Blood Transfusion &             10 &       500 &        500 &          1.0 &             0.2  \\
           Banknote Authentication &             10 &       500 &        500 &          0.8 &             0.6  \\
                           Titanic &             10 &       500 &        250 &          0.2 &             0.3  \\
                           Raisins &             10 &       100 &         50 &          1.0 &             0.6  \\
                              Rice &             10 &       100 &        250 &          0.4 &             0.3  \\
                    Echocardiogram &             10 &        50 &        100 &          0.8 &             0.4  \\
Wisconsin Diagnostic Breast Cancer &             10 &       250 &        100 &          0.2 &             0.7  \\
                        Loan House &             10 &       500 &        500 &          1.0 &             0.2  \\
                     Heart Failure &             10 &       500 &         50 &          0.4 &             0.6  \\
                     Heart Disease &             10 &       400 &        500 &          0.6 &             0.4  \\
                             Adult &             10 &       100 &        500 &          0.6 &             0.6  \\
                    Bank Marketing &             10 &       500 &        250 &          0.8 &             0.6  \\
                   Cervical Cancer &             10 &       100 &        500 &          0.4 &             0.9  \\
              Congressional Voting &             10 &       500 &        500 &          1.0 &             0.7  \\
                       Absenteeism &             10 &       500 &        500 &          1.0 &             0.7  \\
                         Hepatitis &             10 &       500 &        250 &          0.2 &             0.3  \\
                            German &             10 &       250 &        500 &          0.4 &             0.8  \\
                          Mushroom &             10 &       400 &        500 &          0.6 &             0.4  \\
                       Credit Card &             10 &        50 &         50 &          0.4 &             0.1  \\
                       Horse Colic &             10 &        50 &        100 &          0.8 &             0.4  \\
                           Thyroid &             10 &       500 &        500 &          1.0 &             0.7  \\
                          Spambase &             10 &       250 &        500 &          0.8 &             0.6  \\
\midrule
         Iris &             10 &       100 &        500 &          1.0 &             0.5  \\
Balance Scale &             10 &       500 &        250 &          0.6 &             0.4  \\
          Car &             10 &       500 &        500 &          1.0 &             0.7  \\
        Glass &             10 &       500 &        500 &          0.8 &             0.6  \\
Contraceptive &             10 &       250 &        500 &          0.4 &             0.5  \\
  Solar Flare &             10 &       500 &        250 &          0.6 &             0.4  \\
         Wine &             10 &       100 &         50 &          0.2 &             0.8  \\
          Zoo &             10 &       500 &        500 &          1.0 &             0.7  \\
 Lymphography &             10 &       500 &        250 &          0.6 &             0.4  \\
      Segment &             10 &       100 &        500 &          0.4 &             0.9  \\
  Dermatology &             10 &       100 &        500 &          1.0 &             0.5  \\
      Landsat &             10 &       500 &        500 &          1.0 &             0.7  \\
    Annealing &             10 &       100 &        500 &          0.4 &             0.9  \\
       Splice &             10 &       100 &        250 &          0.2 &             0.8  \\
\bottomrule
\end{tabular}
}%
\caption{GeneticTree Hyperparameters}

\label{tab:hpos_gentree}
\end{table}

\begin{table}[H]
\centering
\small
\resizebox{0.85\columnwidth}{!}{
\begin{tabular}{lrllrrr}
\toprule
                      Dataset Name &  max\_depth & criterion & max\_features &  min\_samples\_leaf &  min\_samples\_split &  ccp\_alpha \\
\midrule
                 Blood Transfusion &           9 &   entropy &          None &                   1 &                    5 &         0.0 \\
           Banknote Authentication &           9 &      gini &          None &                   1 &                    5 &         0.0 \\
                           Titanic &           7 &   entropy &          None &                   1 &                   50 &         0.0 \\
                           Raisins &           8 &      gini &          None &                   5 &                    2 &         0.2 \\
                              Rice &           8 &      gini &          None &                   5 &                    2 &         0.2 \\
                    Echocardiogram &           9 &   entropy &          None &                   1 &                    5 &         0.0 \\
Wisconsin Breast Cancer &           7 &   entropy &          None &                   5 &                    2 &         0.4 \\
                        Loan House &          10 &   entropy &          None &                  10 &                    2 &         0.0 \\
                     Heart Failure &           9 &      gini &          None &                   5 &                   50 &         0.0 \\
                     Heart Disease &           8 &   entropy &          None &                   5 &                   10 &         0.0 \\
                             Adult &          10 &   entropy &          None &                  10 &                    2 &         0.0 \\
                    Bank Marketing &           8 &   entropy &          None &                   5 &                   10 &         0.0 \\
                   Cervical Cancer &           9 &   entropy &          None &                   1 &                    5 &         0.0 \\
              Congressional Voting &          10 &      gini &          None &                   1 &                    2 &         0.0 \\
                       Absenteeism &           7 &   entropy &          None &                   1 &                   10 &         0.0 \\
                         Hepatitis &           9 &   entropy &          None &                   1 &                    5 &         0.0 \\
                            German &           7 &   entropy &          None &                   1 &                   10 &         0.0 \\
                          Mushroom &           9 &   entropy &          None &                   1 &                    5 &         0.0 \\
                       Credit Card &           9 &      gini &          None &                   1 &                    5 &         0.0 \\
                       Horse Colic &          10 &   entropy &          None &                  10 &                    2 &         0.0 \\
                           Thyroid &          10 &   entropy &          None &                  10 &                    2 &         0.0 \\
                          Spambase &          10 &      gini &          None &                   1 &                    2 &         0.0 \\
\midrule
         Iris &           8 &   entropy &          None &                   5 &                   10 &         0.0 \\
Balance Scale &           9 &   entropy &          None &                   1 &                    5 &         0.0 \\
          Car &           9 &      gini &          None &                   1 &                    5 &         0.0 \\
        Glass &           9 &      gini &          None &                   1 &                    5 &         0.0 \\
Contraceptive &           9 &   entropy &          None &                   1 &                    5 &         0.0 \\
  Solar Flare &           7 &   entropy &          None &                   1 &                   50 &         0.0 \\
         Wine &           9 &      gini &          None &                   1 &                    5 &         0.0 \\
          Zoo &          10 &      gini &          None &                   1 &                    2 &         0.0 \\
 Lymphography &           7 &   entropy &          None &                   1 &                   10 &         0.0 \\
      Segment &           9 &   entropy &          None &                   1 &                    5 &         0.0 \\
  Dermatology &           9 &      gini &          None &                   1 &                    5 &         0.0 \\
      Landsat &          10 &      gini &          None &                   1 &                    2 &         0.0 \\
    Annealing &           9 &      gini &          None &                   1 &                    5 &         0.0 \\
       Splice &           9 &      gini &          None &                   1 &                    5 &         0.0 \\
\bottomrule
\end{tabular}
}
\caption{CART Hyperparameters}

\label{tab:hpos_sklearn}
\end{table}

\section{Datasets}\label{A:dataset}
The datasets along with their specifications and source are summarized in Table~\ref{tab:data-spec}. 
For all datasets, we performed a standard preprocessing: We applied Leave-one-out encoding to all categorical features. 
Similar to \citet{neural_oblivious}, we further perform a quantile transform, making each feature follows a normal distribution. 
We use a random $80\%/20\%$ train-test split for all datasets. To account for class imbalance, we rebalanced the training data using SMOTE~\citep{chawla2002smote} when the minority class accounts for less than $\frac{25}{c-1} \%$ of the data points, where $c$ is the number of classes.
Since GradTree and DNDT require a validation set for early stopping, we performed another $80\%/20\%$ split on the training data for those approaches. The remainder of the approaches utilize the complete training data. 
For DL8.5 additional preprocessing was necessary since they can only handle binary features. Therefore, we one-hot encoded all categorical features and discretized numeric features by one-hot encoding them using quantile binning with $5$ bins.

\begin{table}[H]
\centering
\resizebox{1\columnwidth}{!}{
\begin{tabular}{L{3.2cm}R{1.7cm}R{1.7cm}R{2.5cm}R{1.7cm}L{9cm}}
\toprule
\multicolumn{1}{L{3.2cm}}{\textbf{Dataset Name}}& \multicolumn{1}{R{1.7cm}}{\textbf{Number of Features}}& \multicolumn{1}{R{1.7cm}}{\textbf{Number of Samples}}& \multicolumn{1}{R{3cm}}{\textbf{Samples by Class}}& \multicolumn{1}{R{1.7cm}}{\textbf{Number of Classes}}& \textbf{Link}\\
\toprule

Blood Transfusion& 4& 748& 178  / 570& 2& \url{https://archive.ics.uci.edu/ml/datasets/Blood+Transfusion+Service+Center}\\
Banknote Authentication& 4& 1372& 610  / 762& 2& \url{https://archive.ics.uci.edu/ml/datasets/banknote+authentication}\\
Titanic& 7& 891& 342 / 549& 2& \url{https://www.kaggle.com/c/titanic}\\
Raisin& 7& 900& 450 /450& 2& \url{https://archive.ics.uci.edu/ml/datasets/Raisin+Dataset}\\
Rice& 7& 3810& 1630 / 2180& 2& \url{https://archive.ics.uci.edu/ml/datasets/Rice+\%28Cammeo+and+Osmancik\%29}\\
Echocardiogram& 8& 132& 107 / 25& 2& \url{https://archive.ics.uci.edu/ml/datasets/echocardiogram}\\
Wisconsin Diagnostic Breast Cancer& 10& 569& 212  / 357& 2& \url{https://archive.ics.uci.edu/ml/datasets/breast+cancer+wisconsin+(diagnostic)}\\
Loan House& 11& 614& 422 / 192& 2& \url{https://www.kaggle.com/code/sazid28/home-loan-prediction/data}\\
Heart Failure& 12& 299& 96 /203& 2& \url{https://archive.ics.uci.edu/ml/datasets/Heart+failure+clinical+records}\\
Heart Disease& 13& 303& 164  / 139& 2& \url{https://archive.ics.uci.edu/ml/datasets/heart+disease}\\
Adult& 14& 32561& 7841 / 24720& 2& \url{https://archive.ics.uci.edu/ml/datasets/adult}\\
Bank Marketing& 14& 45211& 5289  / 39922& 2& \url{https://archive.ics.uci.edu/ml/datasets/bank+marketing}\\
Congressional Voting& 16& 435& 267 / 168& 2& \url{https://archive.ics.uci.edu/ml/datasets/congressional+voting+records}\\
Absenteeism& 18& 740& 279 / 461& 2& \url{https://archive.ics.uci.edu/ml/datasets/Absenteeism+at+work}\\
Hepatitis& 19& 155& 32 / 123& 2& \url{https://archive.ics.uci.edu/ml/datasets/hepatitis}\\
German& 20& 1000& 300 / 700& 2& \url{https://archive.ics.uci.edu/ml/datasets/statlog+(german+credit+data)}\\
Mushrooms& 22& 8124& 4208 /3916& 2& \url{https://archive.ics.uci.edu/ml/datasets/mushroom}\\
Credit Card& 23& 30000& 23364  / 6636 & 2& \url{https://archive.ics.uci.edu/ml/datasets/default+of+credit+card+clients}\\
Horse Colic& 26& 368& 232 / 136& 2& \url{https://archive.ics.uci.edu/ml/datasets/Horse+Colic}\\
Thyroid& 29& 9172& 2401 / 6771& 2& \url{https://archive.ics.uci.edu/ml/datasets/thyroid+disease}\\
Cervical Cancer& 31 & 858& 55 / 803& 2& \url{https://archive.ics.uci.edu/ml/datasets/Cervical+cancer+\%28Risk+Factors\%29}\\
Spambase& 57& 4601& 1813  / 2788& 2& \url{https://archive.ics.uci.edu/ml/datasets/spambase}\\
\midrule
Iris& 4& 150& 50 / 50 / 50& 3& \url{https://archive.ics.uci.edu/ml/datasets/iris}\\
Balance Scale& 4& 625& 49 / 288 / 288& 3& \url{https://archive.ics.uci.edu/ml/datasets/balance+scale}\\
Car& 6& 1728& 384 / 69 / 1210 / 65& 4& \url{https://archive.ics.uci.edu/ml/datasets/car+evaluation}\\
Glass& 9& 214& 70 / 76 / 17 / 13 / 9 / 29& 6& \url{https://archive.ics.uci.edu/ml/datasets/glass+identification}\\
Contraceptive& 9& 1473& 629 / 333 / 511& 3& \url{https://archive.ics.uci.edu/ml/datasets/Contraceptive+Method+Choice}\\
Solar Flare& 10& 1389& 1171 / 141 / 40 / 20 / 9 / 4 / 3 / 1& 8& \url{http://archive.ics.uci.edu/ml/datasets/solar+flare}\\
Wine& 12& 178& 59 / 71 / 48& 3& \url{https://archive.ics.uci.edu/ml/datasets/wine}\\
Zoo& 16& 101& 41 / 20 / 5 / 13 / 4 / 8 /10& 7& \url{https://archive.ics.uci.edu/ml/datasets/zoo}\\
Lymphography& 18& 148& 2 / 81 /61 /4 & 4& \url{https://archive.ics.uci.edu/ml/datasets/Lymphography}\\
Segment& 19& 2310& 330 / 330 / 330 / 330 / 330 / 330 / 330& 7& \url{https://archive.ics.uci.edu/ml/datasets/image+segmentation}\\
Dermatology& 34& 366& 112 / 61 / 72 / 49 / 52 / 20& 6& \url{https://archive.ics.uci.edu/ml/datasets/dermatology}\\
Landsat& 36& 6435& 1533 / 703 / 1358 / 626 / 707 /1508& 6& \url{https://archive.ics.uci.edu/ml/datasets/Statlog+(Landsat+Satellite)}\\
Annealing& 38& 798& 8 / 88 / 608 / 60 / 34& 5& \url{https://archive.ics.uci.edu/ml/datasets/Annealing}\\
Splice& 60& 3190& 767 / 768 / 1655& 3& \url{https://archive.ics.uci.edu/ml/datasets/Molecular+Biology+(Splice-junction+Gene+Sequences)}\\

\bottomrule
\end{tabular}
}
\caption{Dataset Specifications.}

\label{tab:data-spec}
\end{table}

\end{document}